\documentclass[11pt,a4paper]{article}
\usepackage{times,latexsym}
\usepackage{url}
\usepackage[T1]{fontenc}

\usepackage[acceptedWithA]{tacl2018v2}
\usepackage{graphicx} 
\usepackage{subfig} 
\usepackage{booktabs} 
\usepackage{multirow} 
\usepackage{comment} 
\usepackage{amssymb} 

\newcommand{\dkl}{$D_{KL}$}
\newcommand{\method}{amnesic probing}

\newcommand\add[1]{#1}

\usepackage{xspace,mfirstuc,tabulary}

\newif\iftaclinstructions
\taclinstructionsfalse 
\iftaclinstructions
\renewcommand{\confidential}{}
\renewcommand{\anonsubtext}{(No author info supplied here, for consistency with
TACL-submission anonymization requirements)}
\newcommand{\instr}
\fi

\iftaclpubformat 

\else

\fi

\title{Amnesic Probing: Behavioral Explanation with Amnesic Counterfactuals}

\author{Yanai Elazar\textsuperscript{1,2} \, Shauli Ravfogel\textsuperscript{1,2} \, Alon Jacovi\textsuperscript{1}  \, Yoav Goldberg\textsuperscript{1,2}\\
\textsuperscript{1}Computer Science Department, Bar Ilan University \\
\textsuperscript{2}Allen Institute for Artificial Intelligence \\
  {\tt  \{yanaiela,shauli.ravfogel,alonjacovi,yoav.goldberg\}@gmail.com}
  }
  
\date{}

\begin{document}
\maketitle
\begin{abstract}

A growing body of work makes use of \textit{probing} to investigate the working of neural models, often considered black boxes. Recently, an ongoing debate emerged surrounding the limitations of the probing paradigm. In this work, we point out the inability to infer behavioral conclusions from probing results and offer an alternative method that focuses on how the information is being used, rather than on what information is encoded.
Our method, \textit{Amnesic Probing}, follows the intuition that the utility of a property for a given task can be assessed by measuring the influence of a causal intervention that removes it from the representation. 
Equipped with this new analysis tool, we can ask questions that were not possible before, e.g. is part-of-speech information important for word prediction?
We perform a series of analyses on BERT to answer these types of questions.
Our findings demonstrate that conventional probing performance is not correlated to task importance, and we call for increased scrutiny of claims that draw behavioral or causal conclusions from probing results.\footnote{The code is available at: \url{https://github.com/yanaiela/amnesic_probing}}

\end{abstract}

\section{Introduction}
\label{sec:intro}

\begin{figure}[t!]
\centering

\includegraphics[width=0.95\columnwidth]{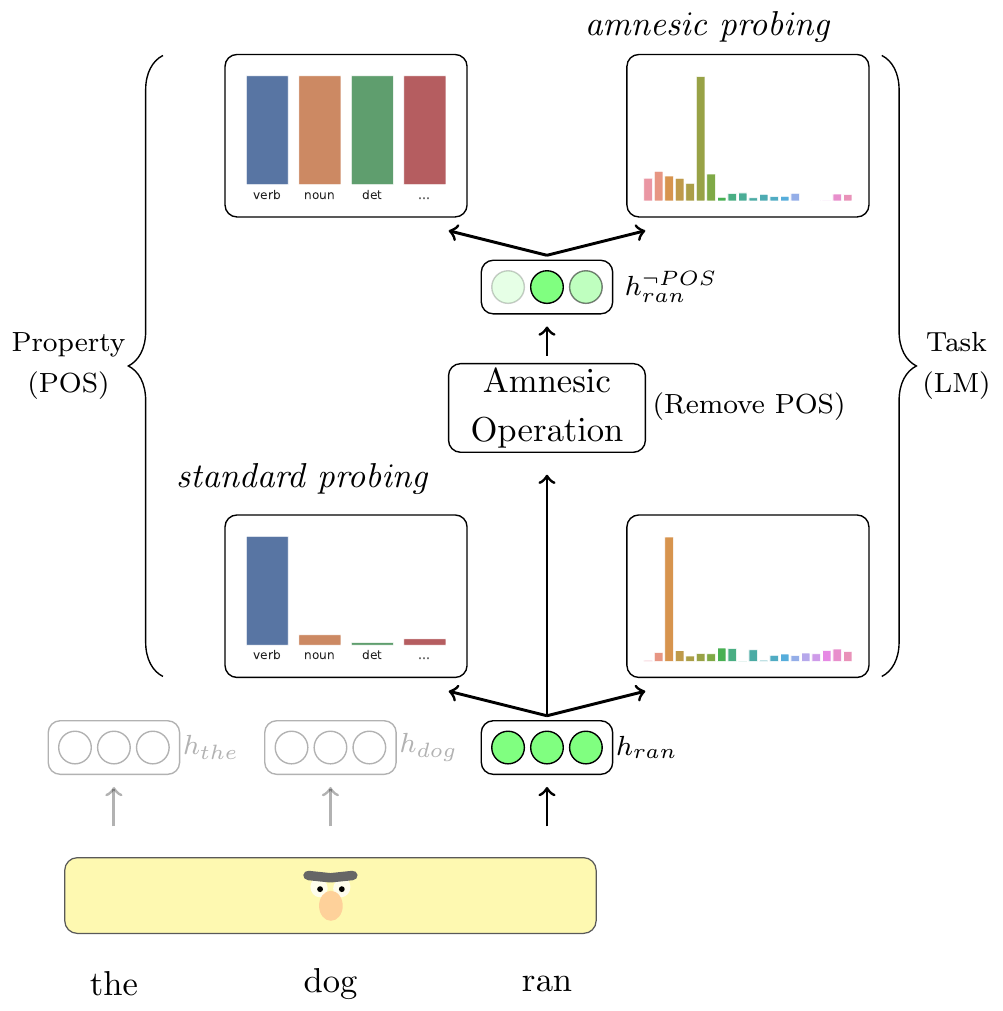}

\caption{A schematic description of the proposed amnesic intervention: we transform the contextualized representation of the word ``ran'' so as to remove information (here, POS), resulting in a ``cleaned'' version $h_{ran}^{\neg POS}$. This representation is fed to the word-prediction layer and the behavioral influence of POS erasure is measured.}
\label{fig:amnestic_probe_overview}
\end{figure}

What drives a model to perform a specific prediction? What information is being used for prediction, and what would have happened if that information went missing?
Since neural representations are opaque and hard to interpret, answering these questions is challenging.

The recent advancements in Language Models (LMs) and their success in transfer learning of many NLP tasks (e.g. \cite{elmo,bert,roberta}) spiked interest in understanding how these models work and what is being encoded in them.
One prominent methodology that attempts to shed light on those questions is probing \cite{cram_vectors_conneau} (also known as \textit{auxilliary prediction} \cite{diagnostic_adi} and \textit{diagnostic classification} \cite{hupkes2018visualisation}). Under this methodology, one trains a simple model -- a \emph{probe} -- to predict some desired information from the latent representations of the pre-trained model. High prediction performance is interpreted as evidence for the information being encoded in the representation. A key drawback of such an approach is that while it may indicate that the information can be extracted from the representation, it provides no evidence for or against the actual use of this information by the model. 
Indeed, \citet{hewitt2019control} have shown that under certain conditions, above-random probing accuracy can be achieved even when the information that is probed for is linguistically-meaningless noise, which is unlikely to have any use by the actual model. 
More recently, \citet{abilasha_probing} showed that models encode linguistic properties, even when not required at all for solving the task, questioning the usefulness and common interpretation of probing.
These results call for higher scrutiny of \emph{causal} claims based on probing results.

In this paper, we propose a counterfactual approach which serves as a step towards causal attribution: \textit{Amnesic Probing} (see Figure \ref{fig:amnestic_probe_overview} for a schematic view). We build on the intuition that if a property $Z$ (e.g., part-of-speech) is being used for a task $T$ (e.g., language-modeling), then the \emph{removal} of $Z$ should negatively influence the ability of the model to solve the task. Conversely, when the removal of $Z$ has little or no influence on the ability to solve $T$, one can argue that knowing $Z$ is not a significant contributing factor in the strategy the model employs in solving $T$. 

As opposed to previous work that focused on intervention in the input space \cite{goyal2019explaining,Kaushik2020Learning,vig2020causal} or in specific neurons \cite{vig2020causal}, our intervention is done on the representation layers. This makes it easier than changing the input (which is non-trivial) and more efficient than querying hundred of neurons (which become combinatorial when considering the effect of multiple neurons simultaneously).

We demonstrate that \method{} can function as a debugging and analysis tool for neural models. Specifically, by using \method{} we show how to deduce whether a property is used by a given model in prediction.

In order to build the counterfactual representations, we need a function that operates on a pre-trained representation and returns a counterfactual version that no longer encodes the property we focus on. We use the recently proposed algorithm for neutralizing linear information: \textit{Iterative Null-space Projection (INLP)} \cite{inlp}.
This approach allows us to ask the counterfactual question: ``How will the prediction of a task differ without access to some property?'' \cite{pearl2018book}.
This approach relies on the assumption that the usefulness of some information can be measured by neutralizing it from the representation and witnessing the resulting \textit{behavioral} change. It echoes the basic idea of ablation tests where one removes some component and measures the influence of that intervention.

We study several linguistic properties such as part-of-speech (POS) and dependency labels.
Overall, we find that as opposed to the common belief, high probing performance does not mean that the probed information is used for predicting the main task (\S \ref{sec:probing_comparison}).
This is consistent with the recent findings of \citet{abilasha_probing}. Our analysis also reveals that the properties we examine are often being used differently in the \textit{masked} setting (which is mostly used in LM training) and in the \textit{non-masked} setting (which is commonly used for probing or fine-tuning) (\S \ref{sec:mlm_deprobing}).
We then dive deeper into a more fine-grained analysis and show that not all of the linguistic property labels equally influence prediction (\S \ref{sec:fine_grained_labels_analysis}). Finally, we re-evaluate previous claims about the way that BERT process the traditional NLP pipeline \cite{nlp_bert-pipeline_tenney} with \method{} and provide a novel interpretation on the utility of different layers (\S \ref{sec:all_layers_deprobing}).

\section{\textit{Amnesic Probing}}
\label{sec:method}

\subsection{Setup and Formulation}

Given a set of labeled data of data points $X=x_1, \dots, x_n$\footnote{The data points can be words, documents, images etc. based on the application.} and task labels $Y=y_1, \dots, y_n$ we analyze a model $f$ that predicts the labels $Y$ from $X$: $\hat{y_i}=f(x_i)$.
We assume that this model is composed of two parts: an encoder $h$ that transforms input $x_i$ into a representation vector $\mathbf{h}_{x_i}$ and a classifier $c$ that is used for predicting $\hat{y_i}$ based on $\mathbf{h}_{x_i}$: $\hat{y_i}=c(h(x_i))$. We refer by \emph{model} to the component that follows the encoding function $h$ and is used for the classification of the task of interest $y$.
Each data point $x_i$ is also associated with a \textit{property} of interest $z_i$ which represents additional information, which may or may not affect the decision of the classifier $c$.

In this work, we are interested in the change in the prediction of the classifier $c$ on the prediction $\hat{y_i}$ which is caused due to the removal of the property $Z$ from the representation $h(x_i)$, that is $h(x_i)^{\neg Z}$.

\subsection{\textit{Amnesic Probing} with INLP}

Under the counterfactual approach, we aim to evaluate the behavioral influence of a specific type of information $Z$ (e.g. POS) on some tasks (e.g. language modeling). To do so, we selectively remove this information from the representation and observe the change in the behavior of the model on the main task.

One commonly used method for information removal relies on adversarial training through the gradient reversal layer technique \cite{ganin2015unsupervised}. However, this technique requires changing the original encoding by retraining the model, which is not desired in our case as we wish to study the original model's behavior. Additionally, \citet{elazar_adversarial} found that this technique does not completely remove all the information from the learned representation.

Instead, we make use of a recently proposed algorithm called Iterative Nullspace Projection (INLP) \cite{inlp}. 
Given a labeled dataset of representations $H$, and a property to remove, $Z$, INLP neutralizes the ability to linearly predict $Z$ from $H$. It does so by training a sequence of linear classifiers (probes) $c_1, ..., c_k$ that predict $Z$, interpreting each one as conveying information on a unique direction in the latent space that corresponds to $Z$, and iteratively removing each of these directions. Concretely, we assume that the $i$th probe $c_i$ is parameterized by a matrix $W_i$. In the $i$th iteration, $c_i$ is trained to predict $Z$ from $H$\footnote{Concretely, we use linear SVM \cite{sklearn}.}, and the data is projected onto its nullspace using a projection matrix $P_{N(W_i)}$. This operation guarantees $W_i P_{N(W_i)}H = 0$, i.e., it neutralizes the features in the latent space which were found by $W_i$ to be indicative to $Z$. By repeating this process until no classifier achieves above-majority accuracy, INLP removes \emph{all} such features.\footnote{All relevant directions are removed to the extent they are identified by the classifiers we train. Therefore, we run INLP until the last linear classifier achieves a score within one point above majority accuracy on the development set.}

\paragraph{Amnesic Probing vs. Probing} 
Note that \method{} \textit{extends} conventional probing, as it is only relevant in cases where the property of interest can be predicted from the representation. 
If a probe gets random accuracy, the information cannot be used by the model, to begin with.
As such, \method{} can be seen as a complementary method, which inspects probe accuracy as a first step, but then proceeds to derive behavioral outcomes from the directions associated with the probe, with respect to a specific task.

\subsection{Controls}
\label{sec:method_controls}
The usage of INLP in this setup involves some subtleties we aim to account for:
(1) Any modification to the representation, regardless of whether it removes information necessary to the task, may cause a decrease in performance. Can the drop in performance be attributed solely to the modification of the representation?
(2) The removal of any property using INLP may also cause the removal of correlating properties. Does the removed information only pertain to the property in question?

\paragraph{Control over Information}

In order to control for the information loss of the representations, we make use of a baseline that removes the same number of directions as INLP does, but randomly.

For every INLP iteration, the data matrix' rank decreases by the number of labels of the inspected property.
This operation removes information from the representation which might be used for prediction.
Using this control, \textit{Rand}, instead of finding the directions using a classifier that learned some task, we generate random vectors from a uniform distribution, that accounts for random directions.
Then, we construct the projection matrix as in INLP, by finding the intersection of nullspaces.

If the \textit{Rand} impact on performance is lower than the impact of \method{} for some property, we conclude that we removed important directions for the main task.
Otherwise, when the \textit{Rand} control has a similar or higher impact, we conclude that there is no evidence for property usage by the main task.

\paragraph{Control over Selectivity\footnote{Not to be confused with \citet{hewitt2019control} Selectivity. Although recommended to use when performing standard probing, we argue it does not fit as a control for \textit{amnesic probing} and provide a detailed explanation in Appendix \ref{sec:hewitt_controls_discussion}.}}

The result of the \method{} is taken as an indication of whether or not the model we query makes use of the inspected property for prediction.
However, the removed features might solely correlate with the property (e.g. word position in the sentence has a nonzero correlation to syntactic function).
To what extent is the information removal process we employ selective to the property in focus?

We test that by explicitly providing the gold information that has been removed from the representation, and finetuning the subsequent layers (while the rest of the network is frozen). Restoring the original performance is taken as evidence that the property we aimed to remove is enough to account for the damage sustained by the amnesic intervention (it may still be the case that the intervention removes unrelated properties; but given the explicitly-provided property information, the model can make up for the damage). However, if the original performance is not restored, this indicates that the intervention removed more information than intended, and this cannot be accounted for by merely explicitly providing the value of the single property we focused on.

Concretely, we concatenate feature vectors of the studied property to the amnesic representations. Those vectors are 32-dimensional and are initialized randomly, with a unique vector for each value of the property of interest. Those are fine-tuned until convergence.  
We note that as the new representation vectors are of a higher dimension than the original ones, we cannot use the original matrix. For an easier learning process, we use the original embedding matrix and concatenate it with a new embedding matrix, randomly initialized, and treat it as the new decision function.

\section{Studying BERT: Experimental Setup}
\label{sec:lm_experiments}

\subsection{Model}
We use our proposed method to investigate BERT \cite{bert},\footnote{Specifically, \textsc{bert-base-uncased} \cite{Wolf2019HuggingFacesTS}.} a popular and competitive masked language model (MLM) which has recently been the subject of many analysis works (e.g., \citet{structural-probe,liu2019linguistic,nlp_bert-pipeline_tenney}). While most probing works focus on the ability to \emph{decode} a certain linguistic property of the input text from the representation, we aim to understand which information is being \emph{used} by it when predicting words from context. For example, we seek to answer questions such as the following: ``Is POS information used by the model in word prediction?'' 
The following experiments focus on language modeling, as a basic and popular task, but our method is more widely applicable.

\subsection{Studied Properties}
\label{subsec:tasks_datasets}

We focus on six tasks of sequence tagging: coarse and fine-grained part-of-speech tags (\textit{c-pos} and \textit{f-pos} respectively); syntactic dependency labels (\textit{dep}); named-entity labels (\textit{ner}); and syntactic constituency boundaries\footnote{Based on the Penn Treebank syntactic definitions.} which mark the beginning and the end of a phrase (\textit{phrase start} and \textit{phrase end} respectively).

We use the training and dev data of the following datasets for each task: English UD Treebank \cite{mcdonald2013ud} for \textit{c-pos}, \textit{f-pos} and \textit{dep}; and English OntoNotes \cite{weischedel2013ontonotes} for \textit{ner}, \textit{phrase start} and \textit{phrase end}. For training, we use 100,000 random tokens from those datasets.

\subsection{Metrics}
We report the following metrics:\\
\textbf{LM accuracy}: Word prediction accuracy.\\
\textbf{Kullback-Leibler Divergence (\dkl{})}:
We calculate the \dkl{} between the distribution of the model over tokens, before and after the amnesic intervention. This measure focuses on the entire distribution, rather than the correct token only. Larger values imply a more significant change.

\begin{table*}[ht]
\centering

\resizebox{0.7\textwidth}{!}{%
\begin{tabular}{ll|rrrrrr}

\toprule
  &       &    \textit{dep} &    \textit{f-pos} &    \textit{c-pos} &    \textit{ner} &  \textit{phrase start} &  \textit{phrase end} \\
\midrule

	\multirow{3}{*}{Properties} & N. dir & 738 &  585 &  264 &  133 &         36 &       22 \\
	& N. classes & 41 &     45 &     12 &     19 &    2 &       2 \\
    & Majority & 11.44 &  13.22 &  31.76 &  86.09 &         59.25 &       58.51 \\

\midrule
\midrule

 Probing & Vanilla         &     76.00 &   89.50 &  92.34 &  93.53 &         85.12 &       83.09 \\

 \midrule

       \multirow{4}{*}{LM-Acc} & Vanilla          &  94.12 &  94.12 &  94.12 &     94.00 &            94.00 &          94.00 \\
& Rand &  12.31 &  56.47 &  89.65 &  92.56 &         93.75 &       93.86 \\
& Selectivity &   73.78 &   92.68 &   97.26 &   96.06 &         96.96 &       96.93 \\
& Amnesic     &   7.05 &  12.31 &  61.92 &  83.14 &         94.21 &       94.32 \\

 \midrule

         \multirow{2}{*}{LM-D$_{KL}$} &  Rand    &   8.11 &   4.61 &   0.36 &   0.08 &          0.01 &        0.01 \\
         & Amnesic           &   8.53 &   7.63 &   3.21 &   1.24 &          0.01 &        0.01 \\

\bottomrule

\end{tabular}
}
  \caption{Property statistics, probing accuracies and the influence of the amnesic intervention on the model's distribution over words. \textit{dep}: dependency edge identity; \textit{f-pos} and \textit{c-pos}: fine-grained and coarse POS tags; \textit{phrase start} and \textit{phrase end}: beginning and end of phrases. \textit{Rand} refers to replacing our INLP-based projection with removal of an equal number of random directions from the representation. 
  The number of iterations per task can be inferred from: $N. dir / N. classes$.
  }

\label{tbl:probing_comparison}
\end{table*}

\section{To Probe or Not to Probe?}
\label{sec:probing_comparison}

By using the probing technique, different linguistic phenomena such as POS, dependency information and NER \cite{nlp_bert-pipeline_tenney,liu2019linguistic,alt2020probing} have been found to be ``easily extractable'' (typically using linear probes).
A naive interpretation of these results may conclude that since information can be easily extracted by the probing model, this information is being used for predictions.
We show that this is not the case. Some properties such as syntactic structure and POS are very informative and are being used in practice to predict words. However, we also find some properties, such as phrase markers, which the model \emph{does not} make use of when predicting tokens, in contrast to what one can naively deduce from probing results.
This finding is in line with a recent work that observed the same behavior \cite{abilasha_probing}.

For each linguistic property, we report the probing accuracy using a linear model, as well as the word prediction accuracy after removing information about that property.
The results are summarized in Table \ref{tbl:probing_comparison}.\footnote{\add{Note that since we use two different datasets, the UD Treebank for \textit{dep}, \textit{f-pos}, \textit{c-pos} and OntoNotes for \textit{ner}, \textit{phrase-start} and \textit{phrase-end}, the Vanilla LM-Acc performance differ between these setups.}} Probing achieves substantially higher performance over majority across all tasks. Moreover, after neutralizing the studied property from the representation, the performance on that task drops to majority (not presented in the table for brevity). Next, we compare the LM performance before and after the projection and observe a major drop for \textit{dep} and \textit{f-pos} information (decrease of 87.0 and 81.8 accuracy points respectively), and a moderate drop for \textit{c-pos} and \textit{ner} information (decrease of 32.2 and 10.8 accuracy points respectively). 
For these tasks, \textit{Rand} performances on \textit{LM-Acc} are lower than the original scores but substantially higher than the Amnesic scores. Recall that the \textit{Rand} experiment is done with respect to the amnesic probing, thus the number of removed dimensions is the same, but each task may differ in the number of dimensions removed. Furthermore, the \dkl{} metric shows the same trend (but in reverse, as a lower value indicates a smaller change).
\add{We also report the selectivity results, where in most experiments the LM performance is restored, indicating amnesic probing works as expected. Note that the \textit{dep} performance is not fully restored, thus some non-related features must have been coupled and removed with the dependency features. We believe that this happens in part, due to the large number of removed directions.\footnote{Since this experiment involves additional fine-tuning and is not entirely comparable to the Vanilla setup (also due to the additional explicit information), we also experiment with concatenating the inspected features and finetuning. This results in an improvement of 3-4 points, above the Vanilla experiment.}}
These results suggest that to a large degree, the damage to LM performance is to be attributed to the specific information we remove, and not to rank-reduction alone. We conclude that dependency information, POS, and NER are important for word prediction.

Interestingly, for \textit{phrase start} and \textit{phrase end} we observe a small \emph{improvement} in accuracy of 0.21 and 0.32 points respectively. 
The performance for the control on these properties is lower, therefore not only these properties are not important for the LM prediction at this part of the model, they slightly harm it.
The last observation is rather surprising as phrase boundaries are coupled to the structure of sentences, and the words that form them. A potential explanation for this phenomenon is that this information is simply not being used at this part of the model, and is rather being processed in an earlier stage.
We further inspect this hypothesis in Section \ref{sec:all_layers_deprobing}. 
Finally, the probe accuracy does not correlate with task importance as measured by our method (Spearman correlation of 8.5, with a p-value of 0.871).

These results strengthen recent works that question the usefulness of probing as an analysis tool \cite{hewitt2019control,abilasha_probing}, but measure it from the usefulness of properties on the main task. We conclude that high probing performance does not entail this information is being used at a later part of the network.

\section{What Properties are Important for the Pre-Training Objective?}
\label{sec:mlm_deprobing}

\begin{table*}[ht]
\centering
\resizebox{0.7\textwidth}{!}{%

\begin{tabular}{ll|rrrrrr}
\toprule
  &       &    \textit{dep} &    \textit{f-pos} &    \textit{c-pos} &    \textit{ner} &  \textit{phrase start} &  \textit{phrase end} \\
\midrule
\multirow{2}{*}{Properties} & N. dir          &  820 &  675 &  240 &  95 &         35 &       52 \\
& N. classes      &   41 &   45 &   12 &  19 &          2 &        2 \\
& Majority        &   11.44 &   13.22 &   31.76 &  86.09 &         59.25 &       58.51 \\

\midrule
\midrule

Probing & Vanilla & 71.19 &  78.32 &   84.40 &  90.68 &         85.53 &       83.21 \\

\midrule

\multirow{4}{*}{LM-Acc} &
Vanilla          &   56.98 &   56.98 &   56.98 &  57.71 &         57.71 &       57.71 \\
& Rand &    4.67 &   24.69 &   54.55 &  56.88 &         57.46 &       57.27 \\
& Selectivity     &   20.46 &   59.51 &   66.49 &  60.35 &         60.97 &       60.80 \\
& Amnesic     &    4.67 &    6.01 &   33.28 &  48.39 &         56.89 &       56.19 \\

\midrule

\multirow{2}{*}{LM-D$_{KL}$} 
& Rand    &    7.77 &    6.10 &    0.45 &   0.10 &          0.02 &        0.04 \\
& Amnesic           &    7.77 &    7.26 &    3.36 &   1.39 &          0.06 &        0.13 \\

\bottomrule
\end{tabular}
}
\caption{Amnesic probing results for the \textit{masked} representations. Properties statistics, word-prediction accuracy and \dkl{} results for the different properties inspected in this work. We report the Vanilla word prediction accuracy and the Amnesic scores, as well as the Rand and 1-Hot controls which shows minimal information loss and high selectivity (except for the \textit{dep} property which all information was removed). The \dkl{} is also reported for all properties in the last rows which show similar trends as the accuracy performance.}

\label{tbl:deprobing_all_tasks_masked}

\end{table*}

Probing studies tend to focus on representations that are used for an end-task (usually the last hidden layer before the classification layer). In the case of MLM models, the words are not masked when encoding them for downstream tasks.

However, these representations are different from those used during the pre-training LM phase (of interest to us), where the input words are masked.
It is therefore unclear if the conclusions drawn from conventional probing also apply to the way that the pre-trained model operates.

From this section on, unless mentioned otherwise, we report our experiments on the masked words. That is, given a sequence of tokens $x_1, \dots, x_i, \dots, x_n$ we encode the representation of each token $x_i$ using its context, as follows: $x_1, \dots, x_{i-1}, [MASK], x_{i+1}, \dots, x_n$. The rest of the tokens remain intact. We feed these input tokens to BERT, and only use the masked representation of each word in its context $h(x_1, \dots, x_{i-1}, [MASK], x_{i+1}, \dots, x_n)_i$. 

\begin{figure}[t!]
\centering

\subfloat{\includegraphics[width=0.5\columnwidth]{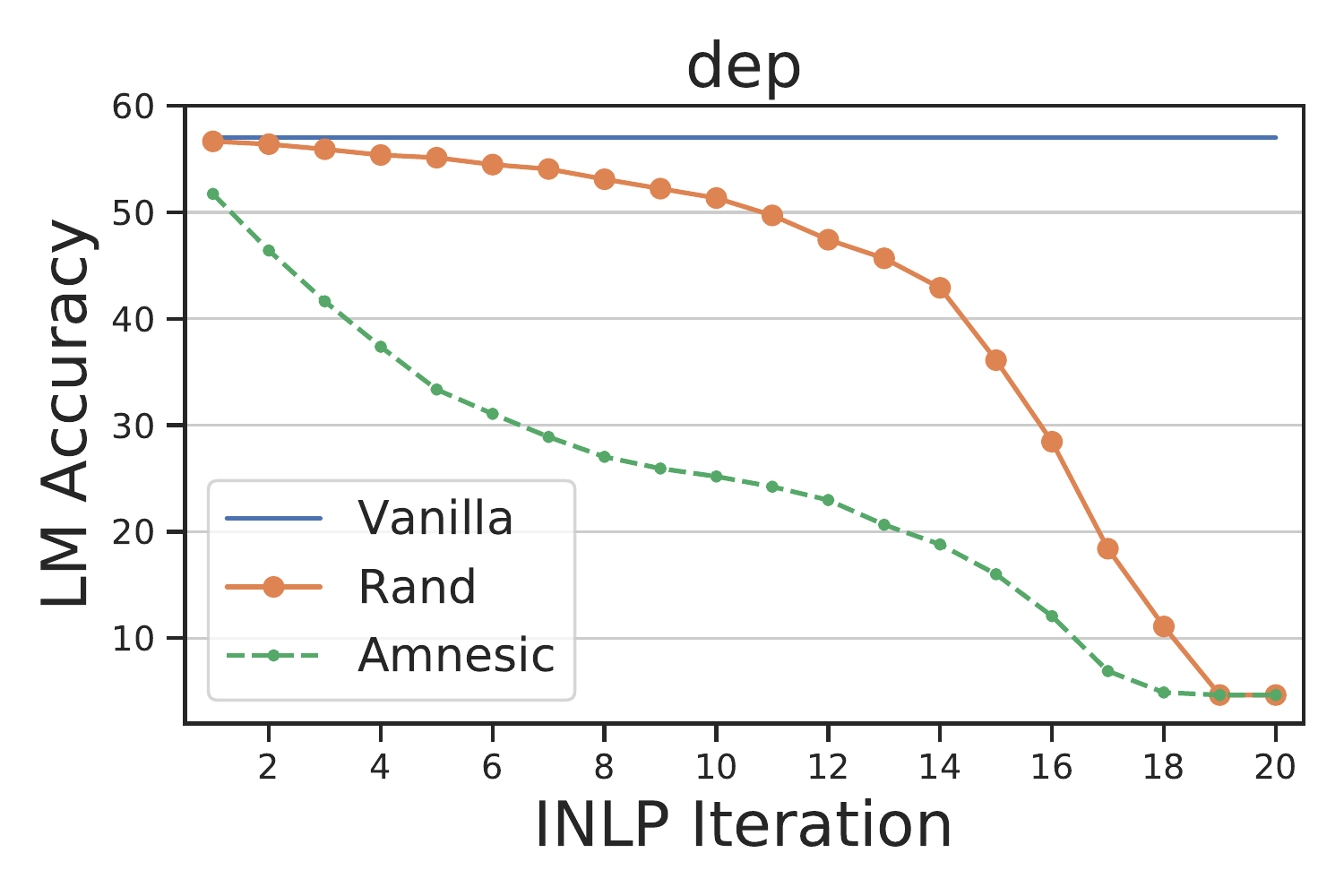}}
\subfloat{\includegraphics[width=0.5\columnwidth]{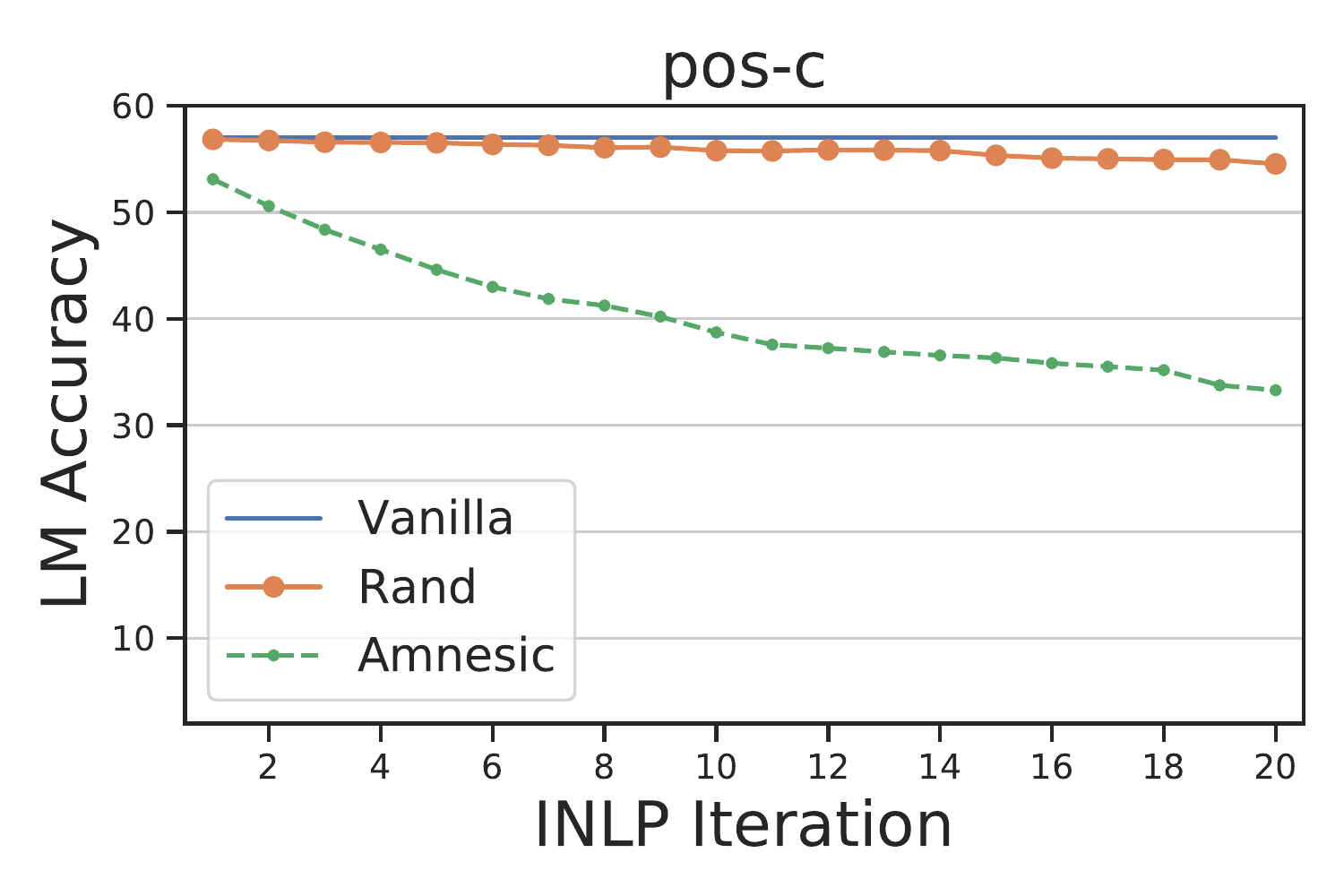}}
\\

\subfloat{\includegraphics[width=0.5\columnwidth]{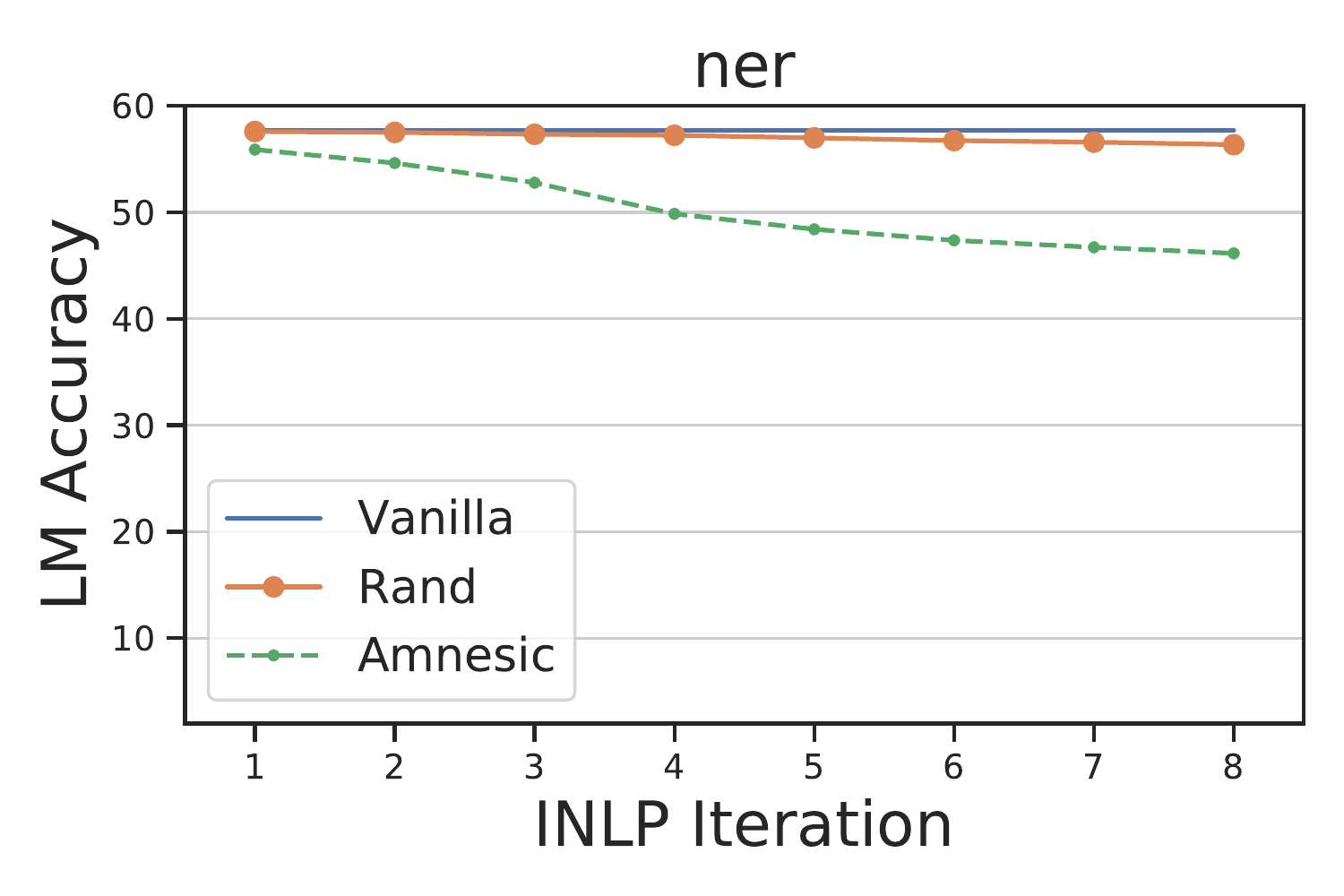}}
\subfloat{\includegraphics[width=0.5\columnwidth]{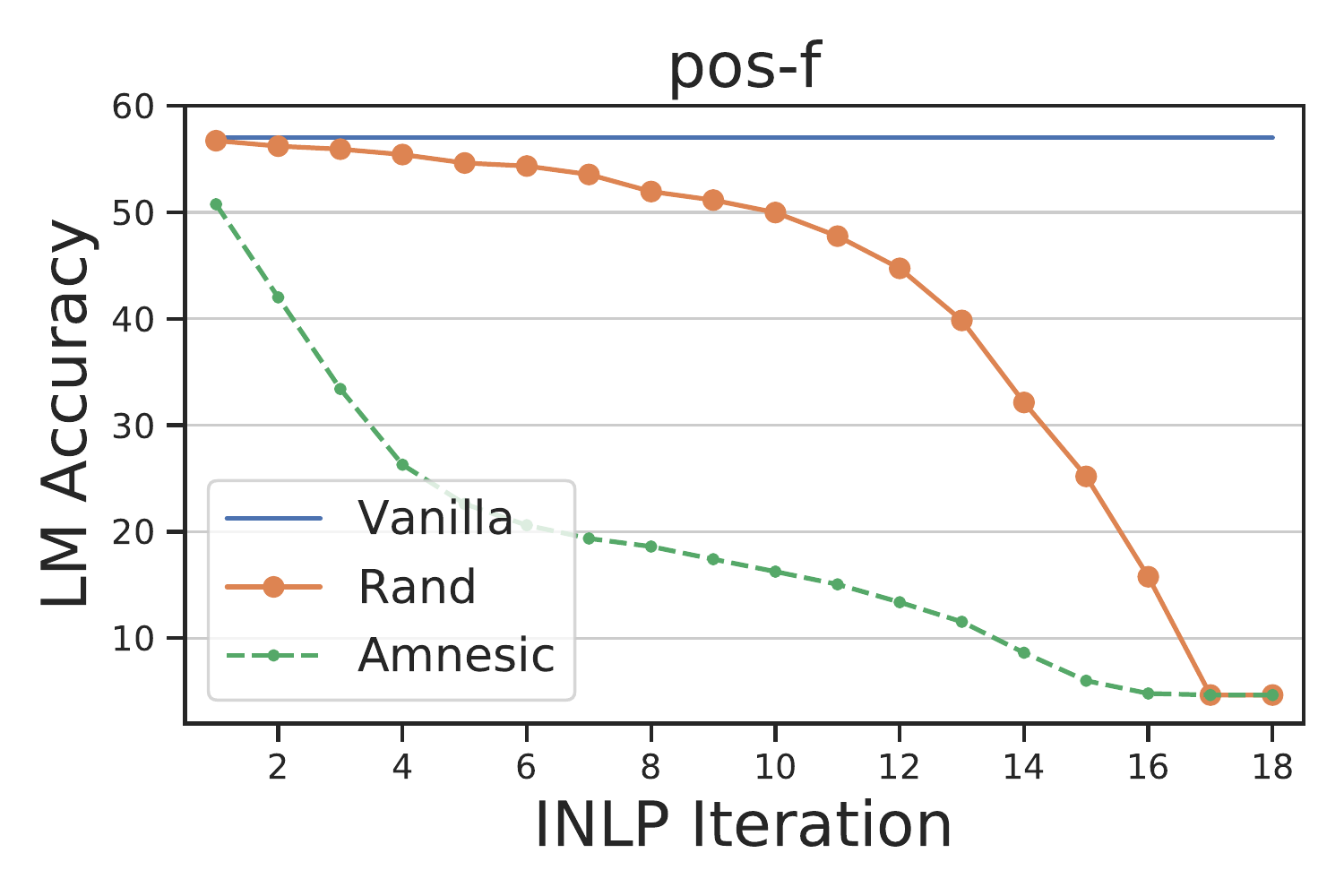}}
\\

\subfloat{\includegraphics[width=0.5\columnwidth]{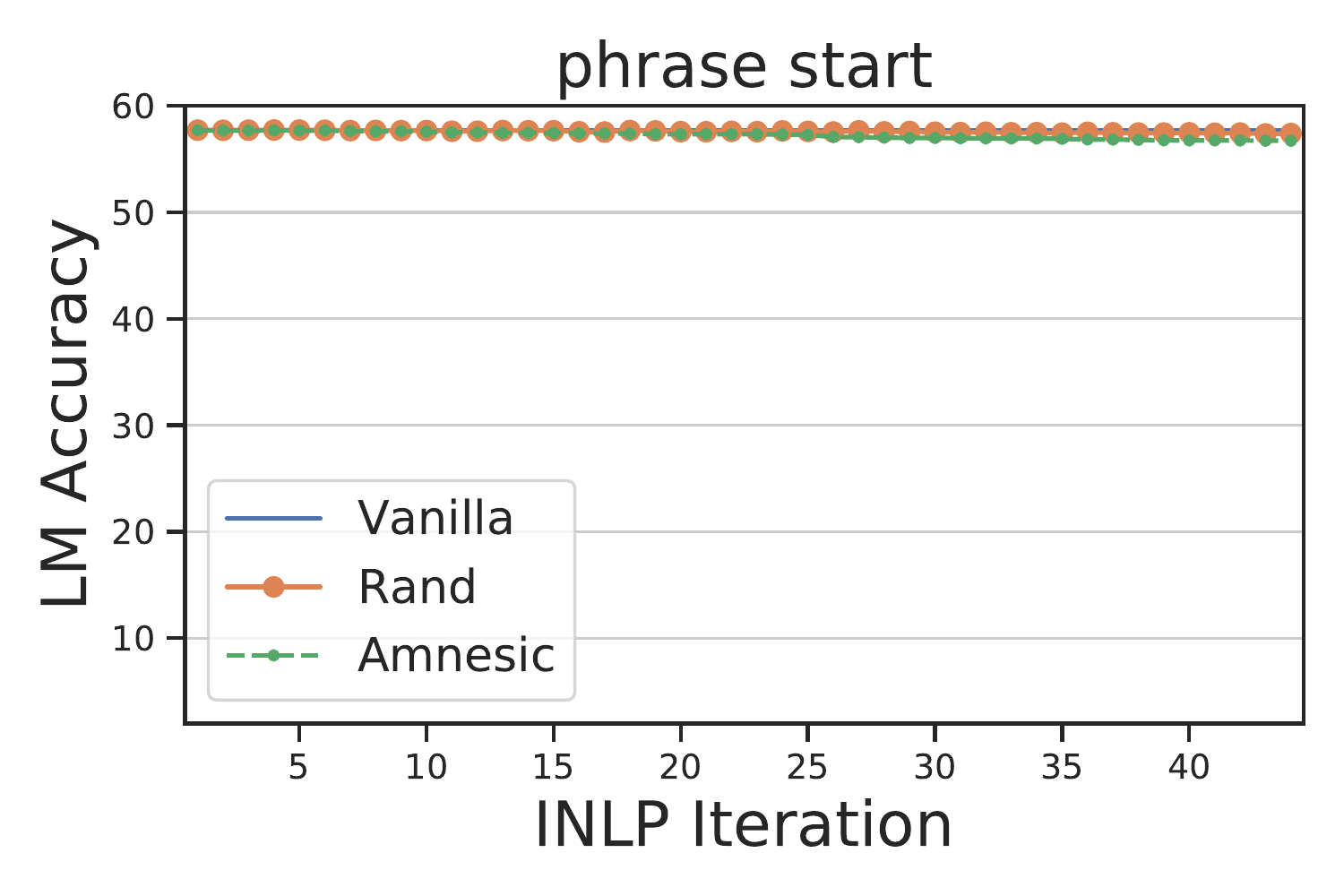}}
\subfloat{\includegraphics[width=0.5\columnwidth]{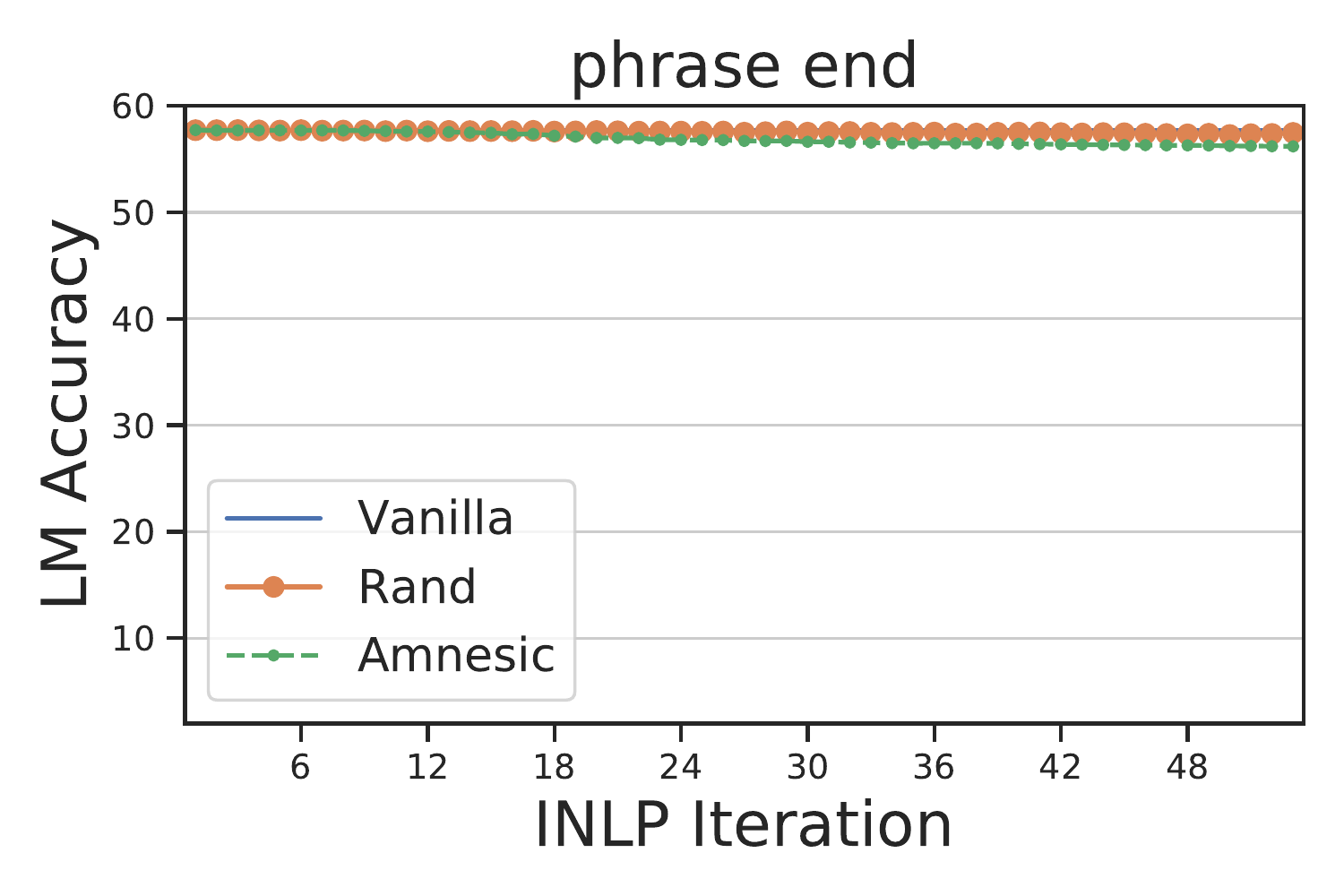}}
\\

\caption{LM accuracy over INLP predictions, for the \textit{masked} tokens version. We present both the Vanilla word-prediction score (straight, blue line), as well as the control (orange, large circles) and INLP (green, small circles). Note that the number of removed dimensions per iteration differs, based on the number of classes of that property.}
\label{fig:INLP_over_time_masked}

\end{figure}

We repeat the experiments from Section \ref{sec:probing_comparison} and report the results in Table \ref{tbl:deprobing_all_tasks_masked}.
As expected, the LM accuracy drops significantly, as the model does not have access to the original word, and it has to infer it only based on context.
Overall, the trends in the \textit{masked} setting are similar to the \textit{non-masked} setting. However, this is not always the case, as we show in Section \ref{sec:all_layers_deprobing}.
We also report the selectivity control. Notice that the performance for this experiment was improved across all tasks. In the case of \textit{dep} and \textit{f-pos}, where we had to neutralize most of the dimensions the performance does not fully recover. 
\add{Note that the number of classes in those experiments might be a factor in the large performance gaps (expressed by the number of removed dimensions, N. dir, in the table). While not part of this study, it would be interesting to control for this factor in future work.}
However for the rest of the properties (\textit{c-pos}, \textit{ner}, and the \textit{phrase-markers}) the performance is fully recovered, showing our methods' selectivity.

To further study the effect of INLP and inspect how the different dimensions removal affects performance, we display in Figure \ref{fig:INLP_over_time_masked} the LM performance after each iteration, both with the \method{} and the control and observe a consistent gap between them.
Moreover, we highlight the difference in the slope for our method and the random direction removal. The \method{} exemplifies a much steeper slope than the random direction, indicating that the studied properties are indeed correlated with word prediction.
\add{We also provide the main task performance after each iteration in Figure \ref{fig:INLP_task_over_time_masked} in the Appendix, which steadily decreases with each iteration.}

\section{Specific Labels and Word Prediction}
\label{sec:fine_grained_labels_analysis}

In the previous sections, we observed the impact (or lack thereof) of different properties on word prediction. But when a property affects word prediction, are all words affected similarly?
In this section, we inspect a more fine-grained version of the properties of interest and study the impact of those on word predictions.

\paragraph{Fine-Grained Analysis}
\begin{table}[t!]
\resizebox{\columnwidth}{!}{%
\centering

\begin{tabular}{lrrrr}
\toprule
                    \textit{c-pos} &  Vanilla & Rand &     Amnesic &  $\Delta$ \\
\midrule
    verb &    46.72 &     44.85 &  34.99 &  11.73 \\
    noun &    42.91 &     38.94 &  34.26 &   8.65 \\
    adposition &    73.80 &     72.21 &  37.86 &  35.93 \\
    determiner &    82.29 &     83.53 &  16.64 &  65.66 \\
    numeral &    40.32 &     40.19 &  33.41 &   6.91 \\
    punctuation &    80.71 &     81.02 &  47.03 &  33.68 \\
    particle &    96.40 &     95.71 &  18.74 &  77.66 \\
    conjunction &    78.01 &     72.94 &   4.28 &  73.73 \\
    adverb &    39.84 &     34.11 &  23.71 &  16.14 \\
    pronoun &    70.29 &     61.93 &  33.23 &  37.06 \\
    adjective &    46.41 &     42.63 &  34.56 &  11.85 \\
    other &    70.59 &     76.47 &  52.94 &  17.65 \\
\bottomrule

\end{tabular}

}
\caption{Masked, \textit{c-pos} removal, fine-grained LM analysis. Removing \textit{c-pos} information and testing the accuracy performance of words, accumulating by their label.
$\Delta$ is the difference in performance between the Vanilla and Amnesic scores.}
\label{tbl:tags_fine_grained_masked_analysis}

\end{table}

When we remove the POS information from the representation, are nouns affected to the same degree as conjunctions?
We repeat the \textit{masked} experimental setting from Section \ref{sec:mlm_deprobing}, but this time we inspect the word prediction performance for the different labels.
We report the results for the \textit{c-pos} tagging in Table \ref{tbl:tags_fine_grained_masked_analysis}.
We observe large differences in the word prediction performance before and after the POS removal between the labels. Nouns, numbers, and verbs show a relatively small impact on performance (8.64, 6.91, and 11.73 respectively), while conjunctions, particles, and determiners demonstrate large performance drops (73.73, 77.66, and 65.65, respectively).
We see that the information about POS labels at the word-level prediction is much more important in closed-set vocabularies (such as conjunctions and determiners) than with open vocabularies (such as nouns and verbs).

\add{A manual inspection of predicted words after removing the POS information reveals that many of the changes are due to the transformation of function words to content words. For example, the words `and', `of', and `a' become `rotate', `say', and `final', respectively, in the inspected sentences. For quantitative analysis, we use a POS tagger in order to measure the POS label confusion before and after the intervention. Out of the 12,700 determiners conjunctions and punctuations, 200 of the predicted words by BERT were tagged as nouns and verbs before the intervention, compared to 3,982 after.}

\begin{figure*}[ht!]
\centering
\subfloat[Non-Masked version]{
\includegraphics[width=0.95\textwidth]{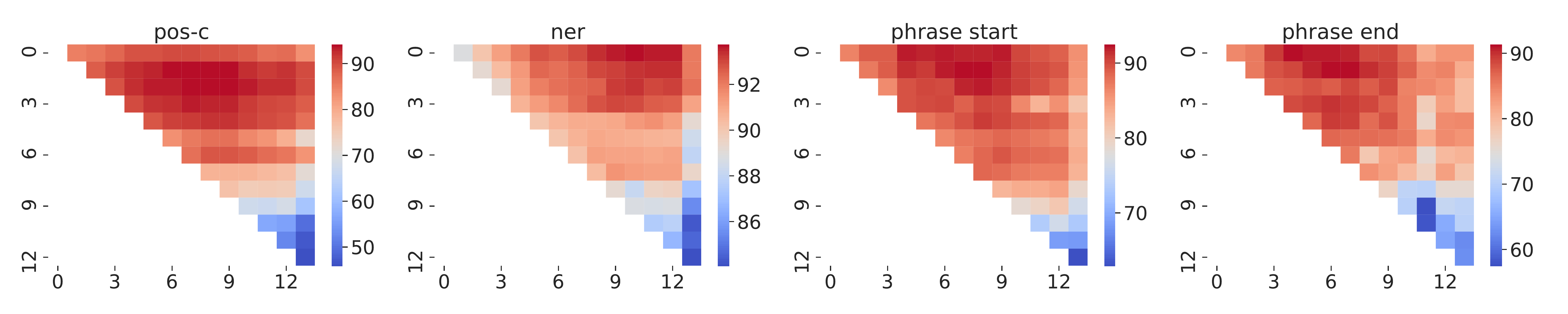}%
\label{fig:layer_wise_deprobe_normal}
}

\subfloat[Masked version]{
\includegraphics[width=0.95\textwidth]{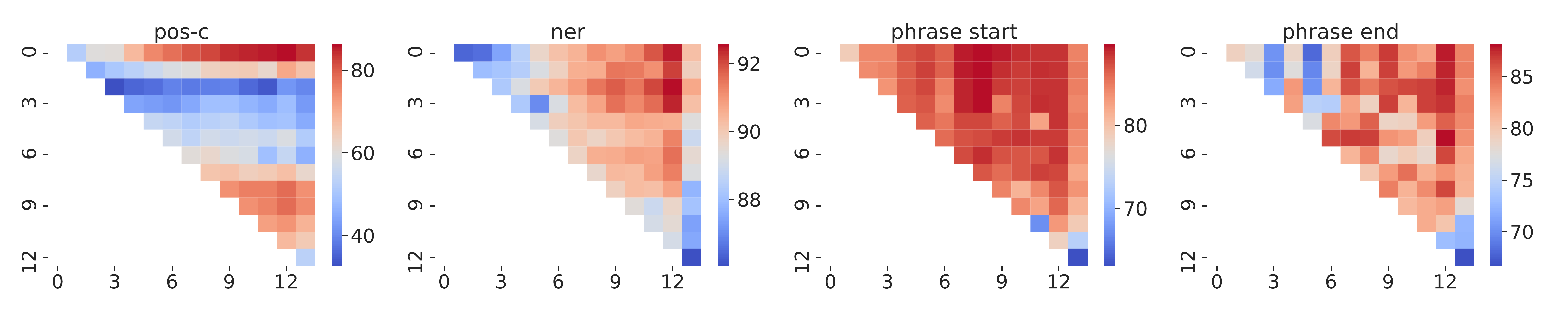}
\label{fig:layer_wise_deprobe_masked}
}

\caption{Layer-wise removal. Removing from layer i (the rows) and testing probing performance on layer j (the columns). Top row (\ref{fig:layer_wise_deprobe_normal}) is non-masked version, bottom row (\ref{fig:layer_wise_deprobe_masked}) is masked.}
\label{fig:layer_wise_removal_probe}
\end{figure*}

\paragraph{Removal of Specific Labels}
\begin{table}[t!]
\centering
\resizebox{0.88\columnwidth}{!}{%

\begin{tabular}{lrrrrr}
\toprule
\textit{c-pos}  &  Vanilla &  Amnesic &  $\Delta$ \\
\midrule
verb        &   56.98 &        55.60 &   1.38 \\
noun        &   56.98 &        55.79 &   1.19 \\
adposition  &   56.98 &        53.40 &   3.58 \\
determiner  &   56.98 &        51.04 &   5.94 \\
numeral     &   56.98 &        55.88 &   1.10 \\
punctuation &   56.98 &        53.12 &   3.86 \\
particle    &   56.98 &        55.26 &   1.72 \\
conjunction &   56.98 &        54.29 &   2.69 \\
adverb      &   56.98 &        55.64 &   1.34 \\
pronoun     &   56.98 &        54.97 &   2.02 \\
adjective   &   56.98 &        55.95 &   1.03 \\
\bottomrule
\end{tabular}

}
\caption{Word prediction accuracy after fine-grained tag distinction removal, \textit{masked} version. Rand control performance are all between 56.05 and 56.49 accuracy (with a maximum difference from Vanilla of 0.92 points).}
\label{tbl:tags_fine_grained_masked}

\end{table}

Following the observation that classes are affected differently when predicting words, we further investigate the differences of specific label removal.
To this end, we repeat the \method{} experiments, but instead of removing the fine-grained information of a linguistic property, we make a cruder removal: the distinction between a specific label and the rest. For example, with POS as the general property, we now investigate whether the information of noun vs. the rest is important for predicting a word.
We perform this experiment for all of the \textit{pos-c} labels, and report the results in Table \ref{tbl:tags_fine_grained_masked}.\footnote{\add{In order to properly compare the different properties, we run INLP for solely 60 iterations, for each property. Since the `other' tag is not common, we omit it from this experiment.}}

We observe big performance gaps when removing different labels.
For example, removing the distinctions between nouns and the rest, or verbs and the rest has minimal impact on performance.
On the other hand, determiners and punctuations are highly affected.
This is consistent with the previous observation on removing specific information. These results call for more detailed observations and experiments when studying a phenomenon as the fine-grained property distinction does not behave the same across labels.\footnote{We repeat these experiments with the other studied properties and observe similar trends.}

\section{Behavior Across Layers}
\label{sec:all_layers_deprobing}

\begin{figure*}[t!]
\centering
\subfloat[Non-Masked version]{
\includegraphics[width=0.95\textwidth]{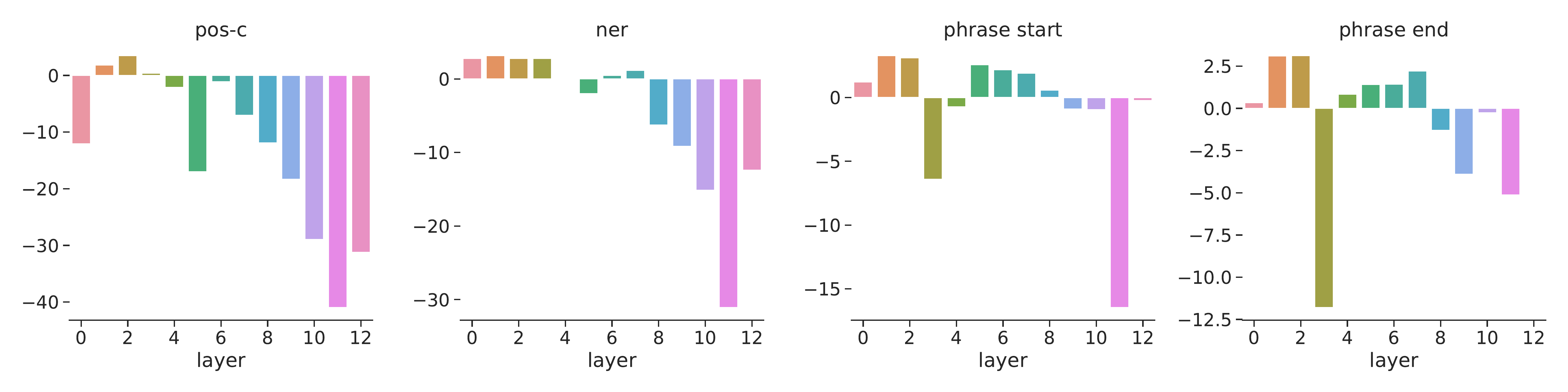}%
\label{fig:layer_influence_unmasked}
}

\subfloat[Masked version]{
\includegraphics[width=0.95\textwidth]{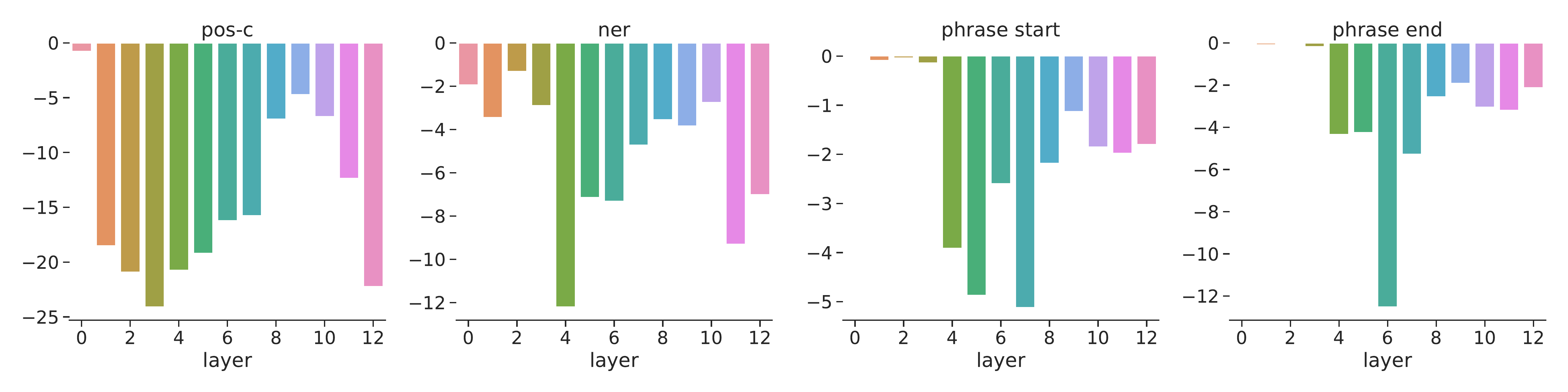}
\label{fig:layer_influence_masked}
}

\caption{The influence of the different properties, from each layer on LM predictions. Top figure (\ref{fig:layer_influence_unmasked}) shows the results on the regular, \textit{non-masked} version, bottom figure (\ref{fig:layer_influence_masked}) for the \textit{masked} version. Colors allow ease of layer comparison across graphs.}
\label{fig:layer_wise_task_lm}
\end{figure*}

The results up to this section treat all of BERT's `Transformer blocks' \cite{transformers} as the encoding function and the embedding matrix as the model. But what happens when we remove the information of some linguistic property from earlier layers?

By using INLP to remove a property from an intermediate layer, we prevent the subsequent layer from using linearly-present information originally stored in that layer.
Though this operation does not erase all the information correlative with the studied property (as INLP only removes linearly-present information), it makes it harder for the model to use this information.
Concretely, we begin by extracting the representation of some text from the first $k$ layers of BERT and then run INLP on these representations to remove the property of interest. 
Given that we wish to study the effect of a property on layer $i$, we project the representation using the corresponding projection matrix $P_i$ that was learned on those representations and then continue the encoding of the following layers.\footnote{As the representations used to train INLP do not include BERTs' special tokens (e.g. `CLS', `SEP'), we also don't use the projection matrix on those tokens.}

\subsection{Property Recovery After an Amnesic Operation}
\label{sec:cross-layer-removal}

Is the property we linearly remove from a given layer recoverable by subsequent layers? We remove the information about some linguistic property from layer $i$ and learn a probe classifier on all subsequent layers $i+1, \dots, n$. This tests how much information about this property the following layers have recovered. 
We experiment with the properties that could be removed without reducing too many dimensions: \textit{pos-c}, \textit{ner}, \textit{phrase start} and \textit{phrase end}. These results are summarized in Figure \ref{fig:layer_wise_removal_probe}, both for the non-masked version (upper row) and the masked version (lower row).

Notably, for the \textit{pos-c}, non-masked version, the information is highly recoverable in subsequent layers when applying the amnesic operation on the first seven layers: the performance drops from the regular probing of that layer between 5.72 and 12.69 accuracy points. However, in the second part of the network, the drop is substantially larger: between 16.57 and 46.39 accuracy points. For the masked version, we witness an opposite trend: The \textit{pos-c} information is much less recoverable in the lower parts of the network than the upper parts.
In particular, the removal of \textit{pos-c} from the second layer appears to affect the rest of the layers, which do not manage to recover a high score on this task, ranging from 32.7 to 42.1 accuracy.

For all of the non-masked experiments, the upper layers seem to make it harder for the subsequent layers to extract the property.
In the masked version, however, there is no consistent trend. It is harder to extract properties after the lower parts for \textit{pos-c} and \textit{ner}. For \textit{phrase start} the upper part makes it harder for further extraction and for \textit{phrase end} both the lower and upper parts make it harder, as opposed to the middle layers. Further research is needed in order to understand the significance of those findings, and whether or not they are related to information usage across layers.

This leads us to the final experiment where we test for the main task performance after an \textit{amnesic} operation at the intermediate layers.

\subsection{Re-rediscovering the NLP Pipeline}

In the previous set of experiments, we measured how much of the signal removed in layer $i$ is recovered in subsequent layers. We now study how the removal of information in layer $i$ affects the word prediction accuracy at the final layer, in order to get a complementary measure for layer importance with respect to a property.
The results for the different properties are presented in Figure \ref{fig:layer_wise_task_lm}, where we plot the difference in word prediction performance between the control and the \method{} when removing a linguistic property from a certain layer.

These results provide a clear interpretation of the internal function of BERT's layers.
For the masked version (Figure \ref{fig:layer_influence_masked}), we observe that the \textit{pos-c} properties are most important in layer 3 and its surrounding layers, as well as layer 12. However, this information is accurately extractable only towards the last layers. For \textit{ner}, we observe that the main performance loss occurs at layer 4. For \textit{phrase-markers} the middle layers are important: layers 5 and 7 for \textit{phrase start} (although the absolute performance loss is not big) and layer 6 for \textit{phrase end} contributes the most for the word prediction performance.

The story with the \textit{non-masked} version is quite different (Figure \ref{fig:layer_influence_unmasked}). First, notice that the amnesic operation \emph{improves} the LM performance for all properties, in some layers.\footnote{\citet{giulianelli2018under} observed a similar behavior by performing an intervention on LSTM activations.} Second, the drop in performance peak across all properties is different than the \textit{masked} version experiments. Particularly, it seems that for \textit{pos-c}, when the words are non-masked in the input, the most important layer for \textit{pos-c} is 11 (and not layer 3, as in the masked version), while this information is easily extractable \add{(by standard probing)} across all layers (above 80\% accuracy).

Interestingly, the conclusions we draw on layer-importance from \method{} partly differ from the ones in the ``Pipeline processing'' hypothesis \cite{nlp_bert-pipeline_tenney}, which aims to localize and attribute information processing of linguistic properties to parts of BERT (for the non-masked version).\footnote{We note that this work analyzes BERT-base, in contrast to \citet{nlp_bert-pipeline_tenney} which analyzed BERT-Large.}
On one hand, the \textit{ner} experiment trends are similar: the last layers are much more important than earlier ones (in particular, layer 11 is the most affected in our case, with a decrease of 31.09 accuracy points.
On the other hand, in contrast to their hypotheses, we find that POS information, \textit{pos-c} (which was considered to be more important in the earlier layers) affects the word prediction performance much more in the upper layers (40.99 accuracy loss in the 11th layer).
Finally, we note that our approach performs an ablation of these properties in the representation space, which reveals which layers are actually responsible for processing properties, as opposed to \citet{nlp_bert-pipeline_tenney} which focused on where this information is easily extractable.

We note the big differences in behavior when analyzing the masked vs. the non-masked version of BERT, and call for future work to make a clearer distinction between the two.
Finally, we stress that the different experiments should not be compared between one setting to the other, and thus the different y-scales in the figures. This is due to confounding variables (e.g. the number of removed dimensions from the representations), which we do not control for in this work.

\section{Related Work}

With the established impressive performance of large pre-trained language models \cite{bert,roberta}, based on the Transformer architecture \cite{transformers}, a large body of work started studying and gaining insight into how these models work and what do they encode.\footnote{
These works cover a wide variety of topics, 
including grammatical generalization \cite{syntax_goldberg2019,warstadt2019investigating}, syntax \cite{tenney2018what,lin2019open,reif2019visualizing,structural-probe,liu2019linguistic}, world knowledge \cite{lama,jiang2019can}, reasoning \cite{talmor2019olmpics}, and commonsense \cite{forbes2019neural,zhou2019evaluating,weir2020existence}.}
For a thorough summary of these advancements we refer the reader to a recent primer on the subject \cite{rogers2020primer}.

A particularly popular and easy-to-use interpretation method is probing \cite{cram_vectors_conneau}.
Despite its popularity, recent works have questioned the use of probing as an interpretation tool. 
\citet{hewitt2019control} have emphasized the need to distinguish between decoding and learning the probing tasks. They introduced \textit{control tasks}, a consistent but linguistically meaningless attribution of labels to tokens, and have shown that probes trained on the control tasks often perform well, due to the strong lexical information held in the representations and learned by the probe. This leads them to propose a selectivity measure that aims to choose probes that achieve high accuracy only on linguistically-meaningful tasks. 
\citet{tamkin2020investigating} claim that probing cannot serve as an explanation of downstream task success.
They observe that the probing scores do not correlate with the transfer scores achieved by fine-tuning. 
 
Finally, \citet{abilasha_probing} show that probing can achieve non-trivial results for linguistic properties that were not needed for the task the model was trained on.
In this work, we observe a similar phenomenon, but from a different angle. We actively remove some property of interest from the queried representation and measure the impact of the \emph{amnesic} representation of the property on the main task.

Two recent works study the probing paradigm from an information-theory perspective. \citet{pimentel2020informationtheoretic} emphasize that under a mutual-information maximization objective, ``better'' probes are increasingly more accurate, regardless of complexity. They use the data-processing inequality to question the rationale behind methods that focus on encoding and propose \emph{ease of extractability} as an alternative criterion. \citet{voita2020information} follow this direction, using the concept of minimum description length \cite[MDL,][]{rissanen1978modeling} to quantify the total information needed to transmit both the probing model and the labels it predicts. Our discussion here is somewhat orthogonal to those on the meaning of encoding and probe complexity, as we focus on the information influence on the model's behavior, rather than on the ability to extract it from the representation.

Finally and concurrent to this work, \citet{causaLM} have studied a similar question of a \emph{causal} attribution of concepts to representations, using adversarial training guided by causal graphs.

\section{Discussion}
\label{sec:discussion}

Intuitively, we would like to completely neutralize the abstract property we are interested in --- e.g., POS information (\emph{completeness}), as represented by the model --- while keeping the rest of the representation intact (\emph{selectivity}).
This is a nontrivial goal, as it is not clear whether neural models actually have abstract and disentangled representations of properties such as POS, which are independent of other properties of the text. It may be the case that the representation of many properties is intertwined. Indeed, there is an ongoing debate on the assertion that certain information is ``encoded'' in the representation \cite{voita2020information, pimentel2020informationtheoretic}. However, even if a disentangled representation of the information we focus on exists, it is not clear how to detect it.

We implement the information removal operation with INLP, which gives a first-order approximation using linear classifiers; we note, however, that one can in principle use other approaches to achieve the same goal. While we show that we do remove the linear ability to predict the properties and provide some evidence to the selectivity of this method (\S \ref{sec:method}), one has to bear in mind that we remove only linearly-present information, and that the classifiers can rely on arbitrary features that happen to correlate with the gold label, be it a result of spurious correlations or inherent encoding of the direct property. \add{Indeed, we observe this behavior in Section \ref{sec:cross-layer-removal} (Figure \ref{fig:layer_wise_removal_probe}), where we neutralize the information from certain layers, but occasionally observe higher probing accuracy in following layers.} We thus stress that the information we remove in practice should be seen only as an approximation for the abstract information we are interested in and that one has to be cautious of \emph{causal} interpretations of the results.
\add{Although in this paper we use the INLP algorithm in order to remove linear information, \textit{\method{}} is not restricted to removing linear information. 
When non-linear removal methods become available, they can be swapped instead of INLP.
This stresses the importance of creating algorithms for non-linear information removal.}

Another unanswered question is how to quantify the \emph{relative} importance of different properties encoded in the representation for the word prediction task. The different erasure portion for different properties makes it hard to draw conclusions on which property is more important for the task of interest. While we do not make claims such as ``dependency information is more important than POS'', these are interesting questions that should be further discussed and researched.

\section{Conclusions}

In this work, we propose a new method, \textit{Amnesic Probing}, which aims to quantify the influence of specific properties on a model that is trained on a task of interest. We demonstrate that conventional probing falls short in answering such behavioral questions, and perform a series of experiments on different linguistic phenomena, quantifying their influence on the masked language modeling task. Furthermore, we inspect both unmasked and masked BERT's representation and detail the differences between them, which we find to be substantial. We also highlight the different influences of specific fine-grained properties (e.g. nouns, and determiners) on the final task. Finally, we use our proposed method on the different layers of BERT, and study which parts of the model make use of the different properties. Taken together, we argue that compared with probing, counterfactual intervention --- such as the one we present here --- can provide a richer and more refined view of the way symbolic linguistic information is encoded and used by neural models with distributed representations.\footnote{All of the experiments were logged and tracked using Weights and Biases \cite{wandb}.}


\section*{Acknowledgments}
We would like to thank Hila Gonen, Amit Moryossef, Divyansh Kaushik, Abhilasha Ravichander, Uri Shalit, Felix Kreuk Jurica \v{S}eva and Yonatan Belinkov for their helpful comments and discussions.
We also thank the anonymous reviewers and the action editor, Radu Florian, for their valuable suggestions.

This project has received funding from the Europoean Research Council (ERC) under the Europoean Union's Horizon 2020 research and innovation programme, grant agreement No. 802774 (iEXTRACT).
Yanai Elazar is grateful to be partially supported by the PBC fellowship for outstanding Phd candidates in Data Science.

\bibliography{tacl2018}
\bibliographystyle{acl_natbib}

\newpage
\clearpage

\appendix

\section{Appendix}

\begin{figure}[h]
\centering

\subfloat{\includegraphics[width=0.5\columnwidth]{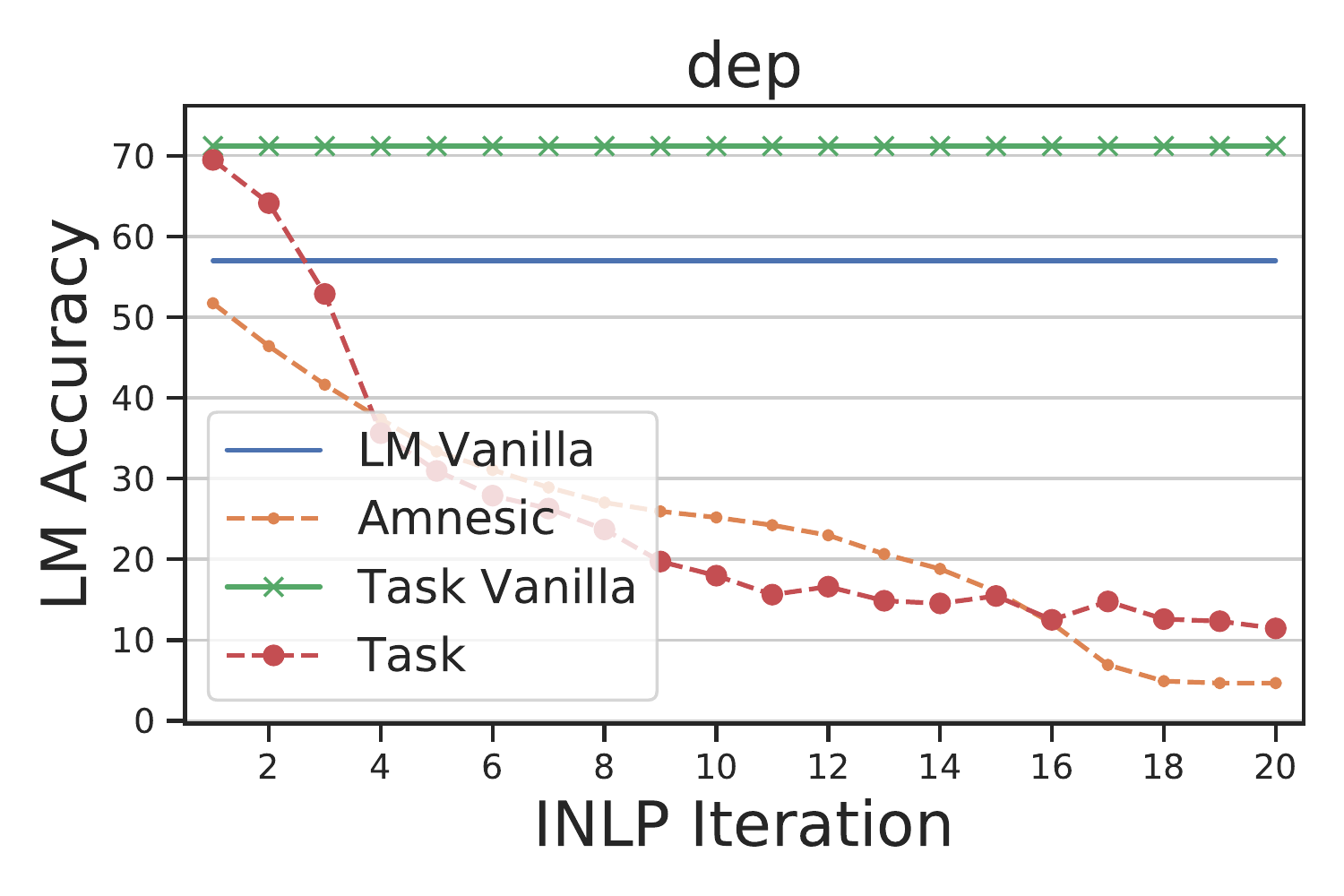}}
\subfloat{\includegraphics[width=0.5\columnwidth]{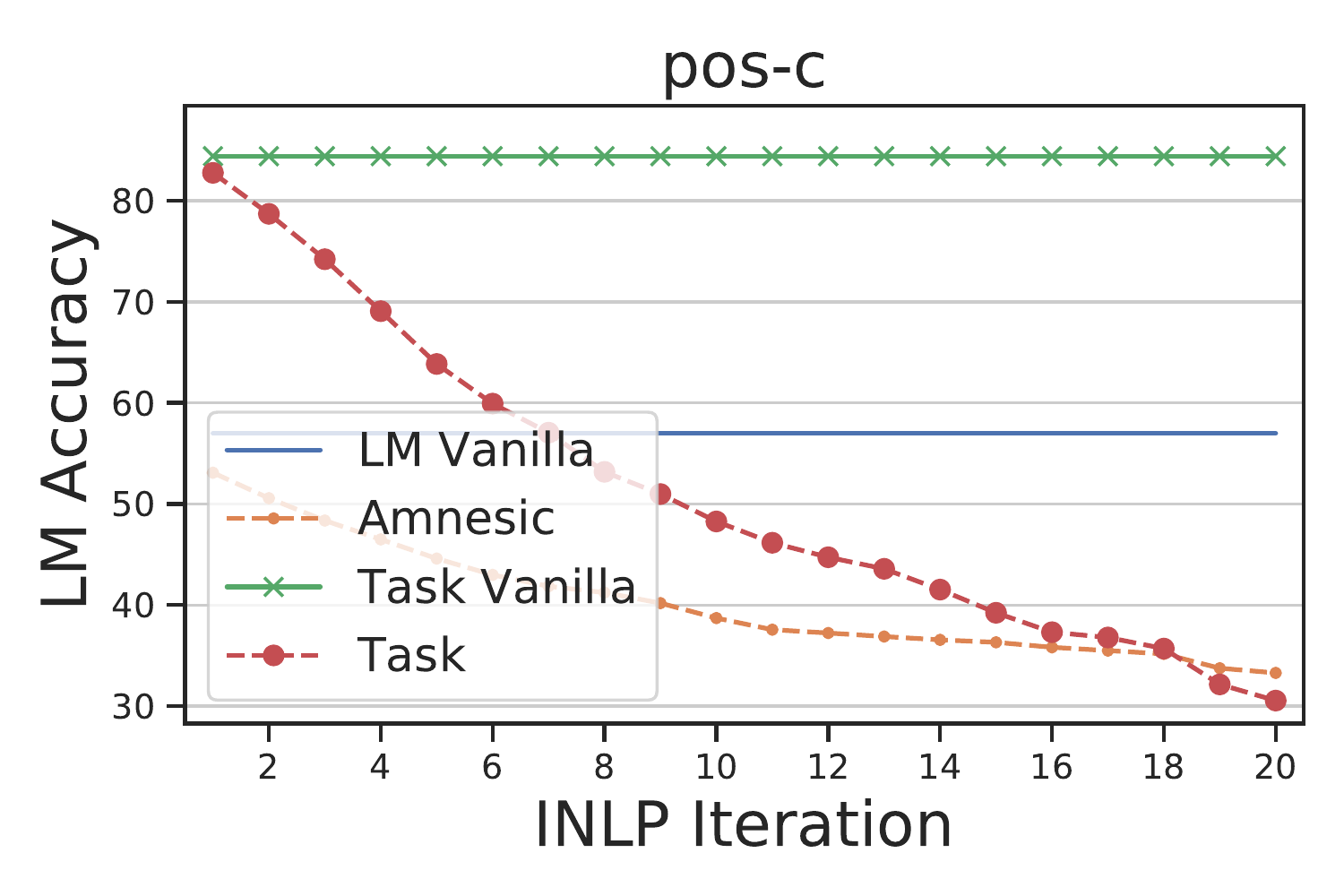}}
\\

\subfloat{\includegraphics[width=0.5\columnwidth]{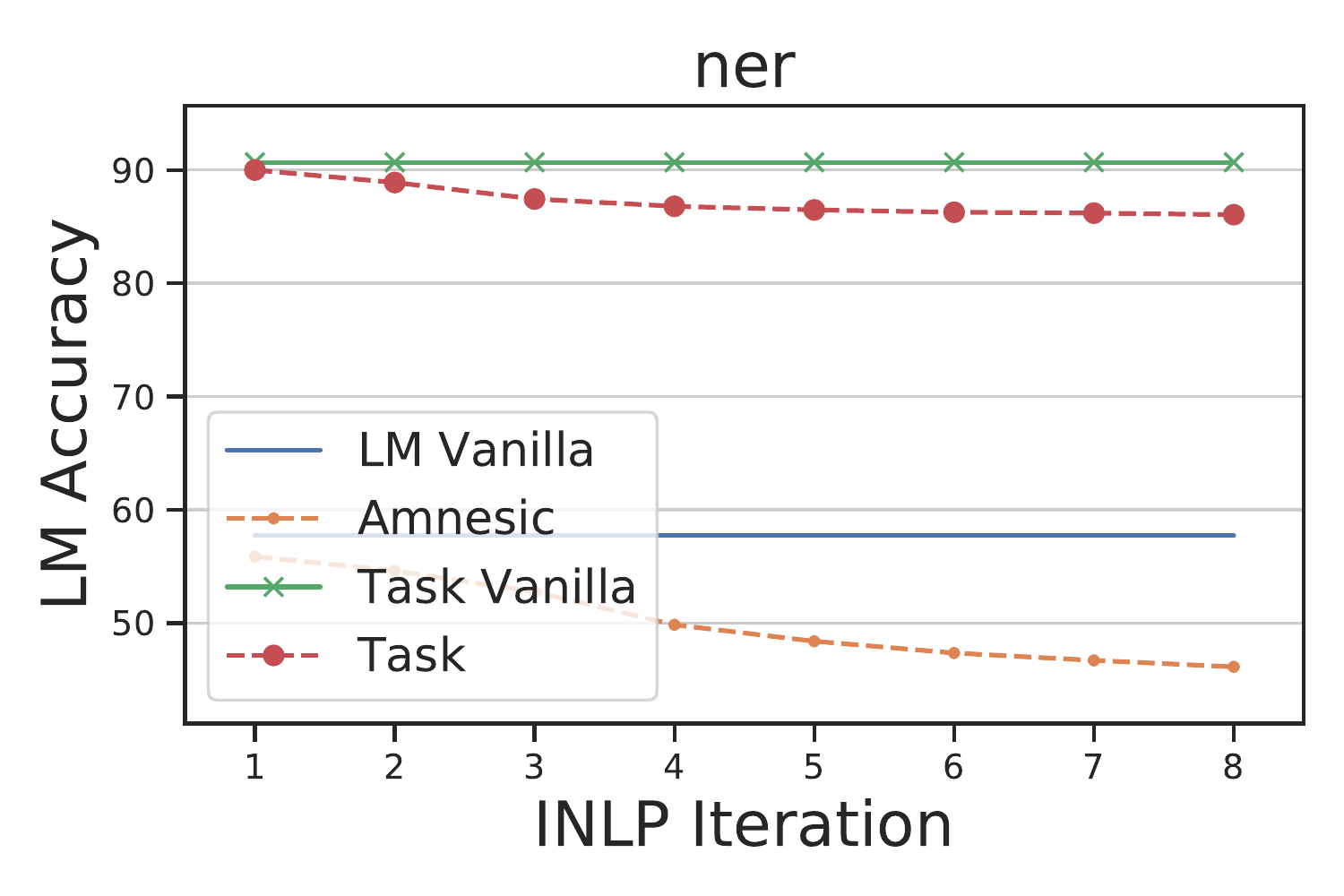}}
\subfloat{\includegraphics[width=0.5\columnwidth]{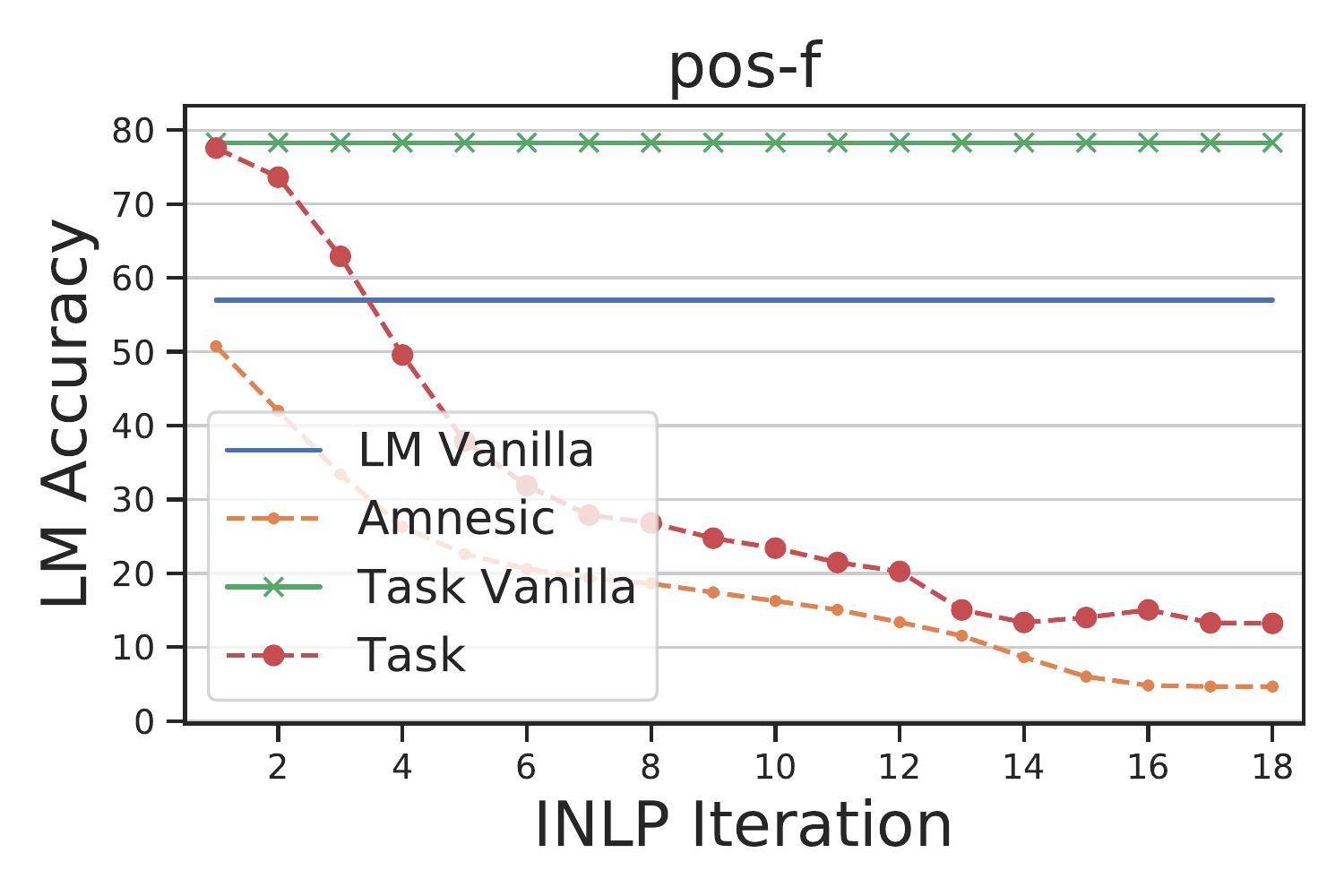}}
\\

\subfloat{\includegraphics[width=0.5\columnwidth]{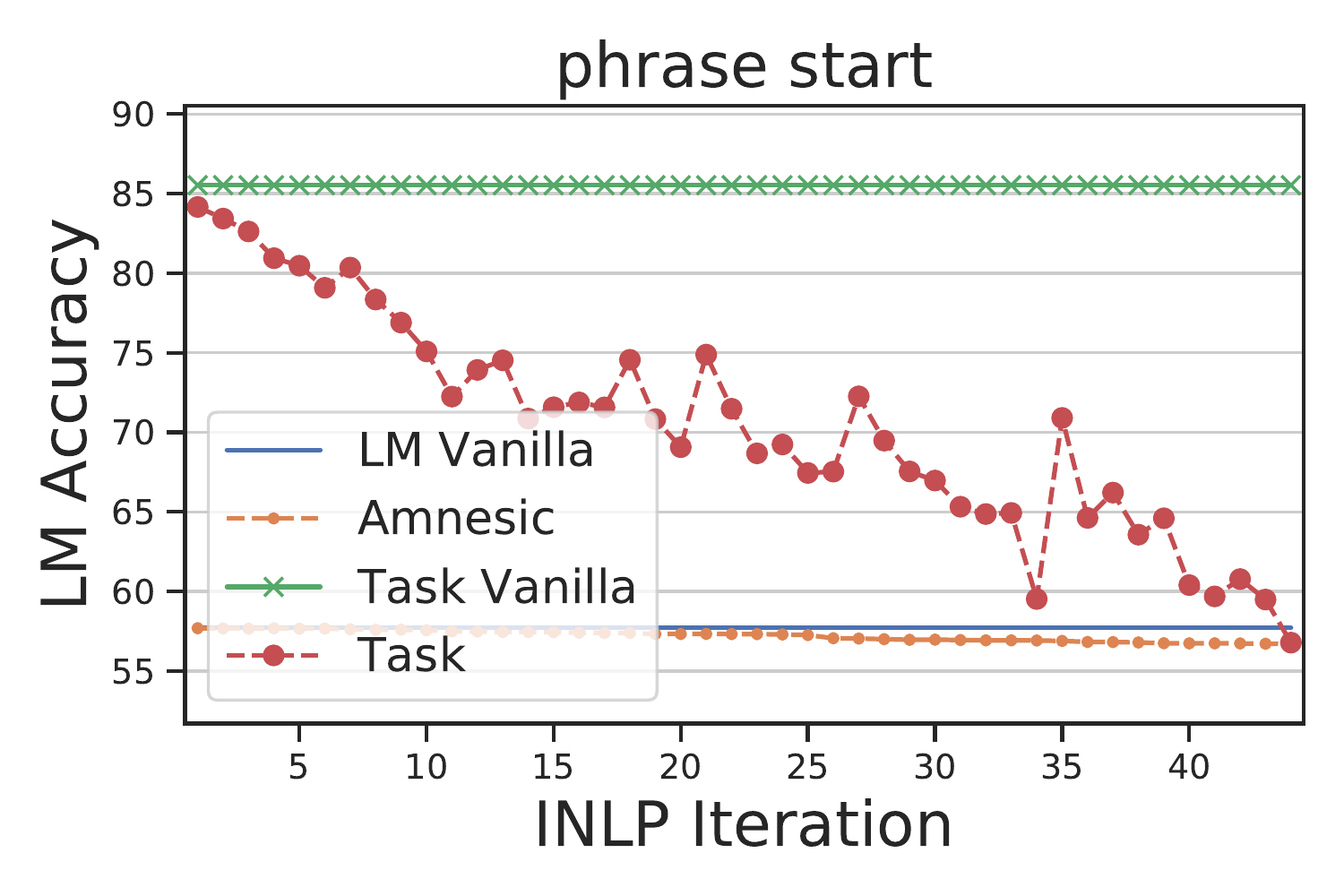}}
\subfloat{\includegraphics[width=0.5\columnwidth]{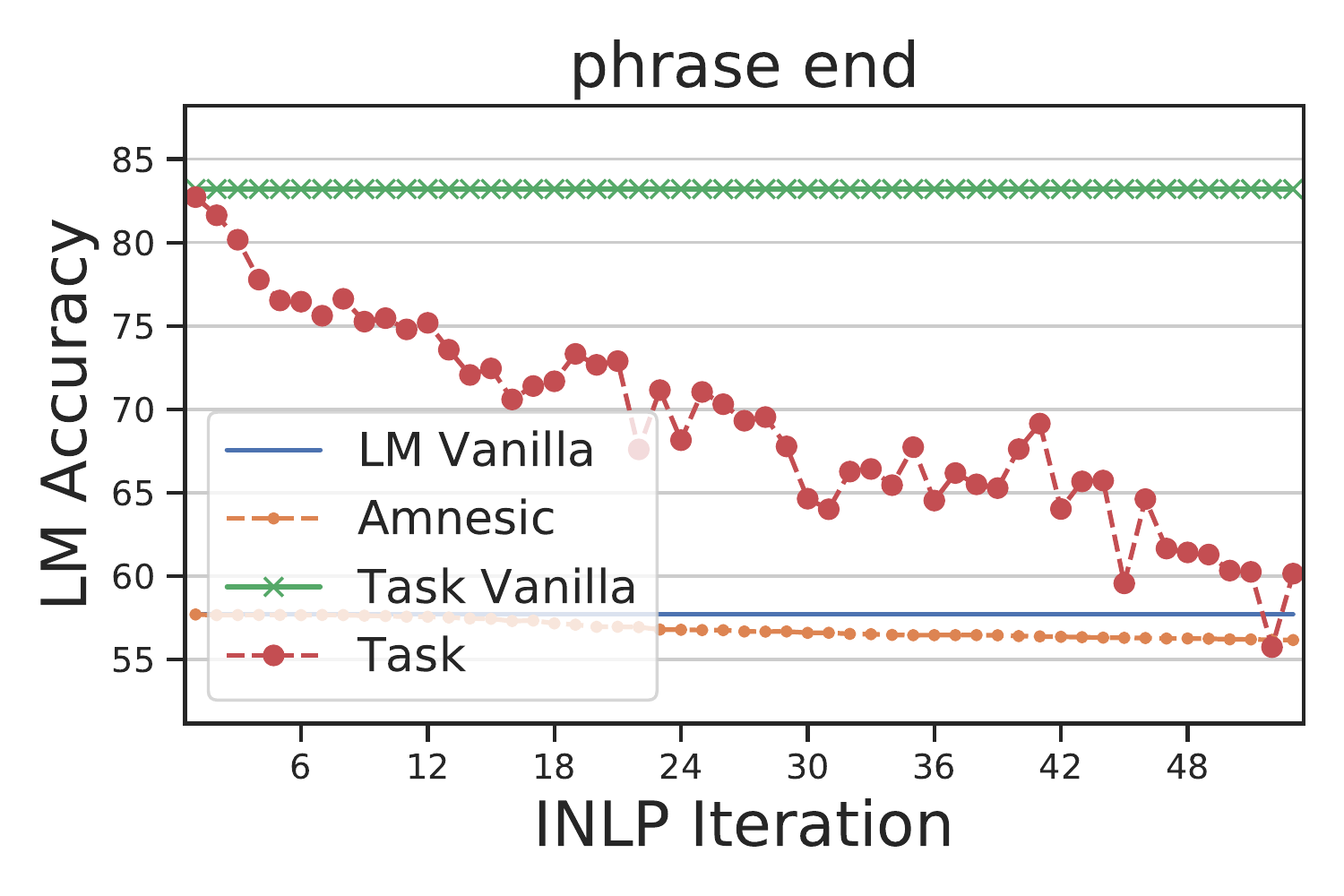}}
\\

\caption{LM accuracy over INLP predictions, for the \textit{masked} tokens version. We present both the Vanilla word-prediction score (straight, blue line), as well as Amnesic Probing (orange, small circles), and the main task performance (red, large circles). For reference, we also provide the vanilla probing performance of each task (green, cross marks). Note that the number of removed dimensions per iteration differs, based on the number of classes of that property.}
\label{fig:INLP_task_over_time_masked}

\end{figure}

We provide additional experiments that depict the performance of the main task (e.g. POS) performance during the INLP iterations in Figure \ref{fig:INLP_task_over_time_masked}.

\section{Hewitt and Liang's Control Task}
\label{sec:hewitt_controls_discussion}
\begin{figure}[t!]
\centering

\subfloat{\includegraphics[width=0.5\columnwidth]{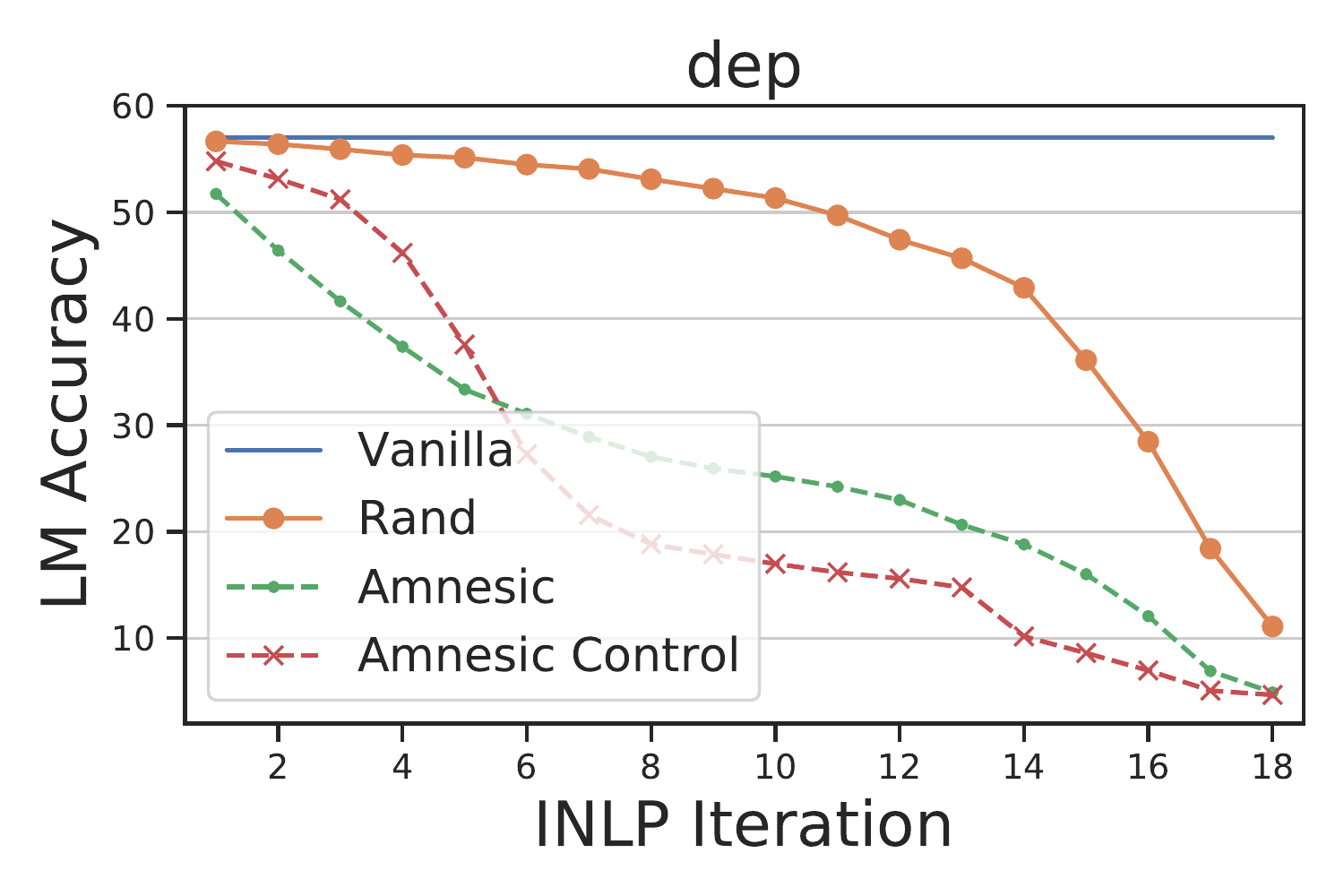}}
\subfloat{\includegraphics[width=0.5\columnwidth]{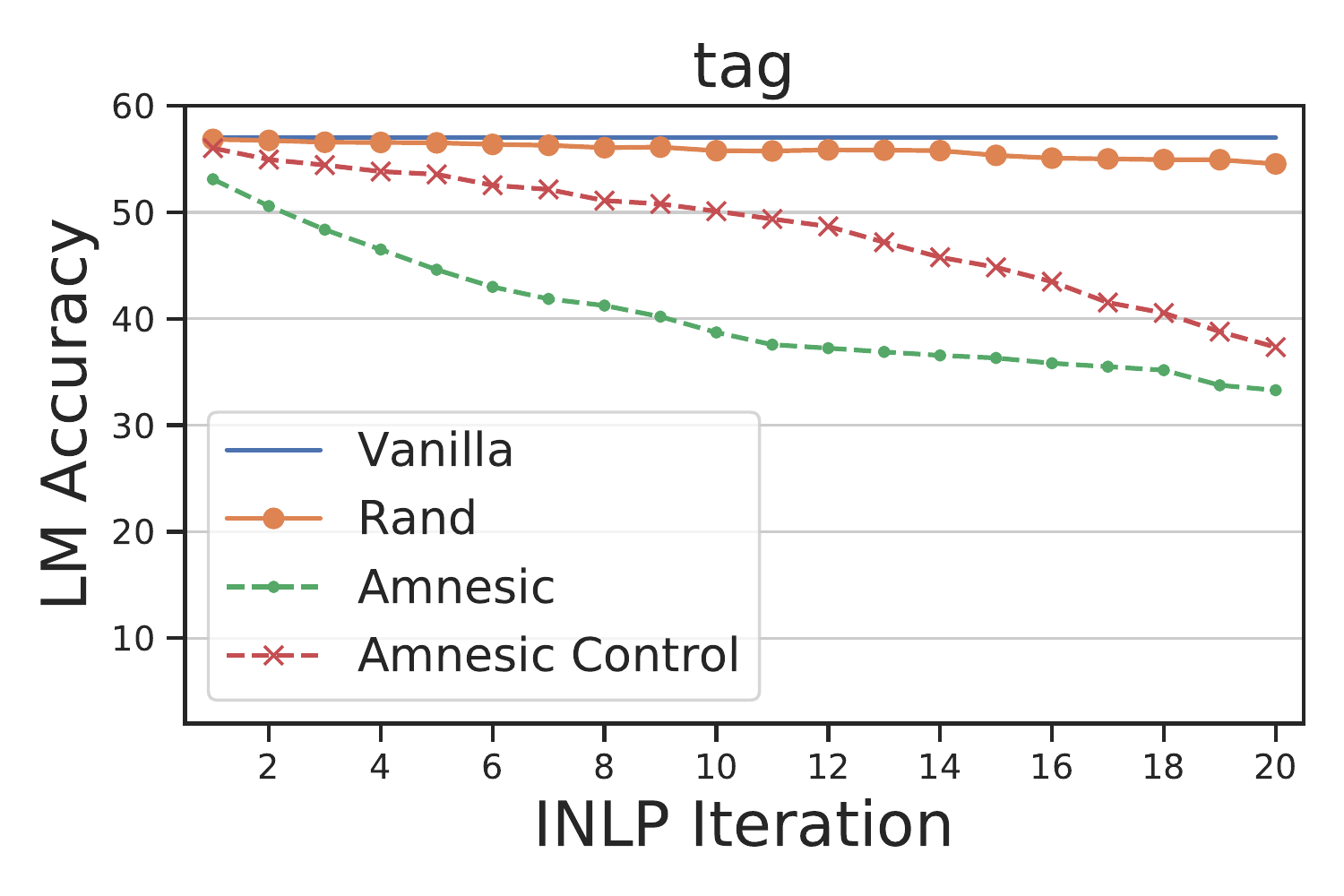}}
\\

\subfloat{\includegraphics[width=0.5\columnwidth]{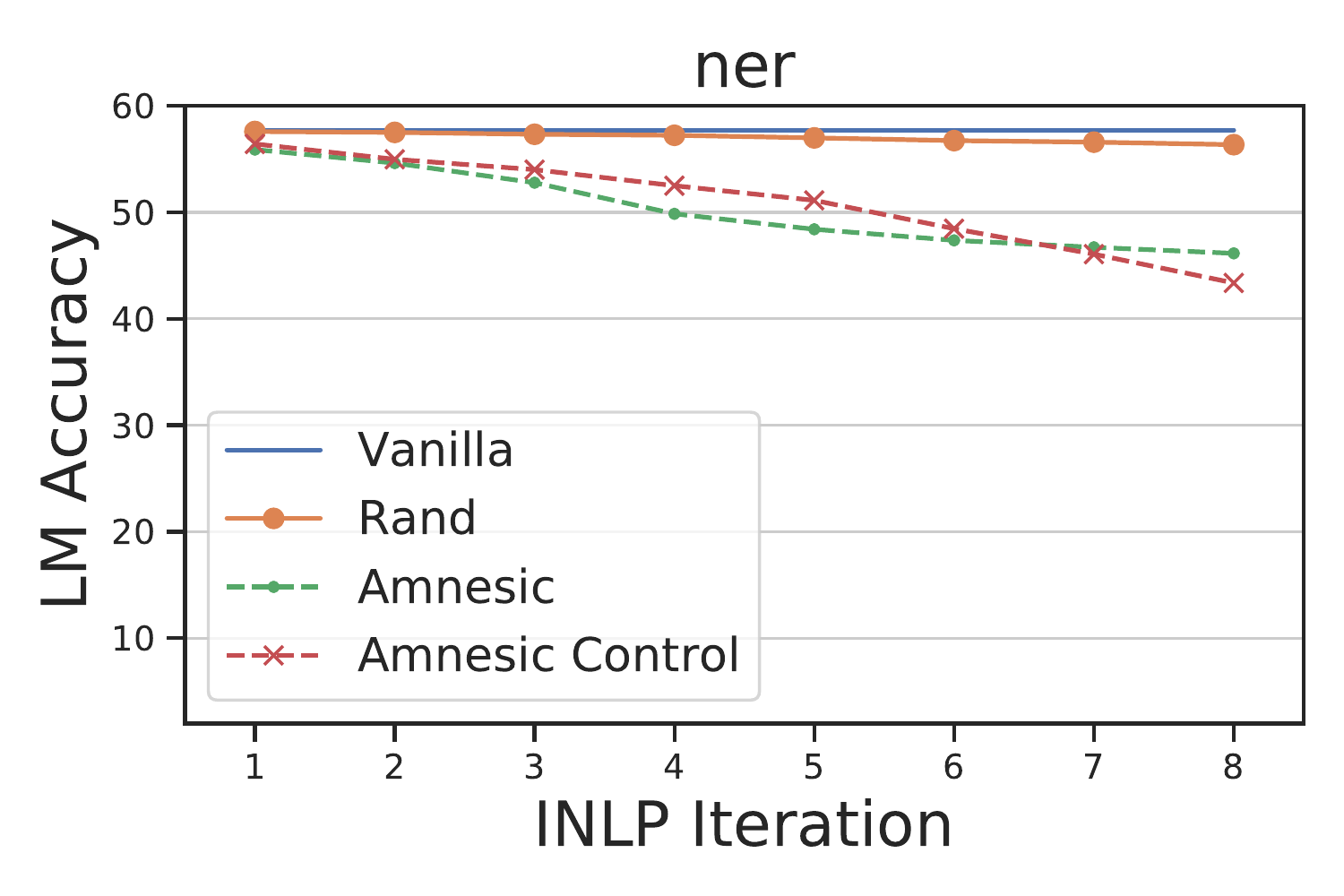}}
\subfloat{\includegraphics[width=0.5\columnwidth]{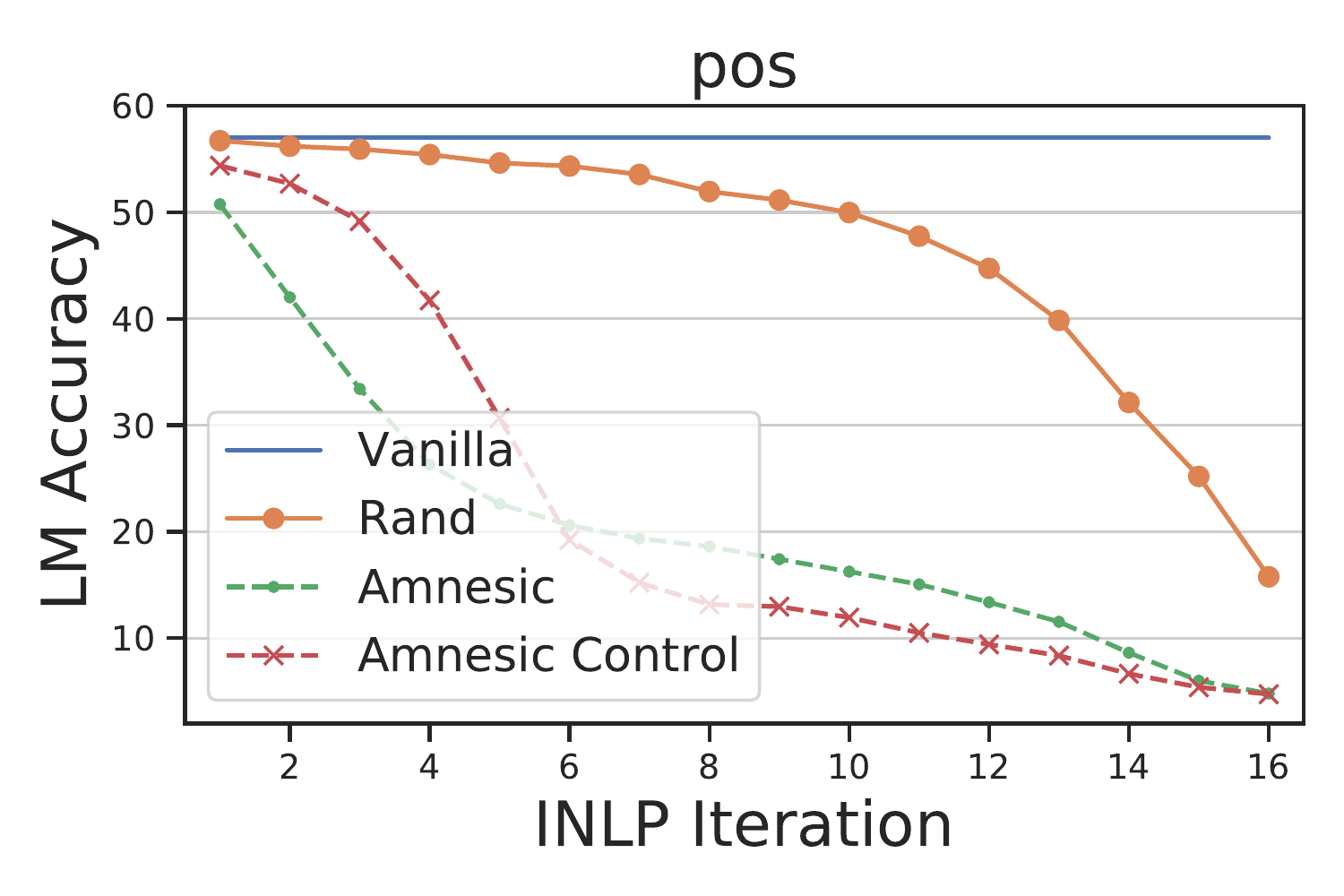}}
\\

\subfloat{\includegraphics[width=0.5\columnwidth]{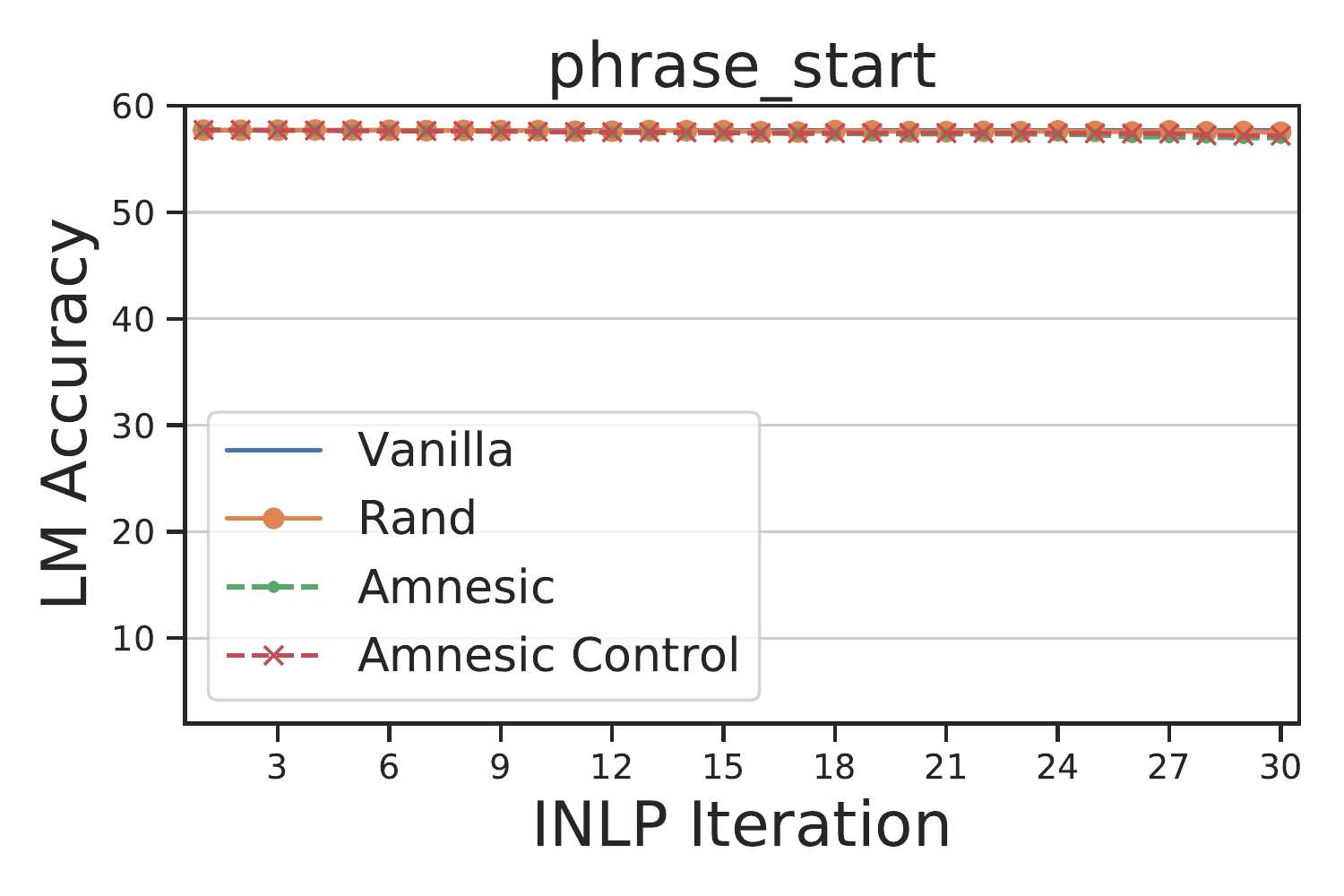}}
\subfloat{\includegraphics[width=0.5\columnwidth]{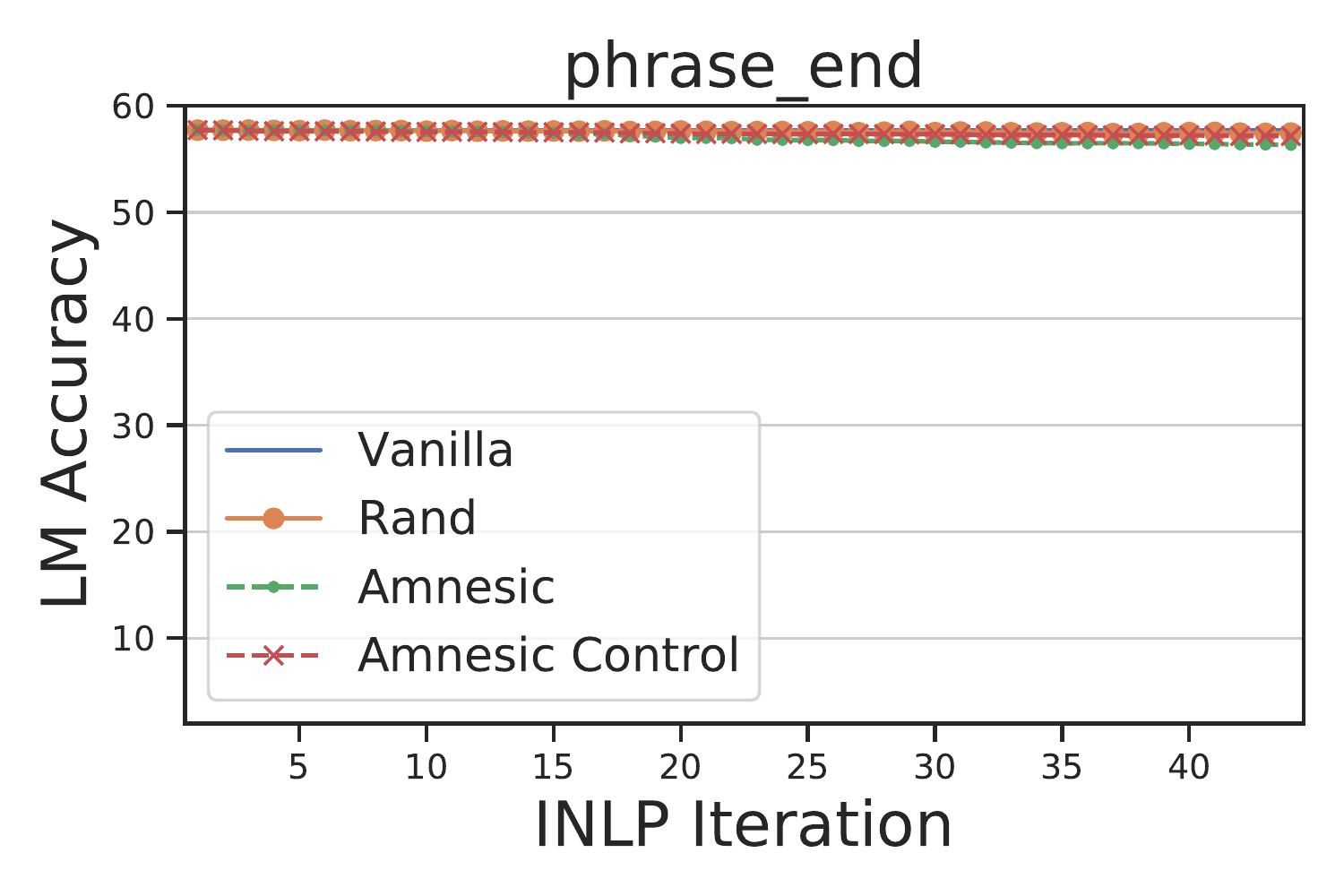}}
\\

\caption{LM accuracy over INLP predictions, for the \textit{masked} tokens version. We present both the Vanilla word-prediction score (straight, blue line), as well as Amnesic Probing (orange, small circles), and the control performance (orange, large circles). We also provide the Control results for the selectivity test, proposed by \citet{hewitt2019control} (red, crosses). Note that the number of removed dimensions per iteration differs, based on the number of classes of that property.}
\label{fig:INLP_task_over_time_masked_hewitt_control}

\end{figure}

\textit{Control Task} \cite{hewitt2019control} has been suggested as a way to attribute the performance of the probe to the extraction of \textit{encoded} information, as opposed to lexical memorization. Our goal in this work, however, is not to extract information from the representation (as is done in conventional probing) but to measure a behavioral outcome. Since the control task is solved by lexical memorization, applying INLP on control task's classifiers erases lexical information (i.e., erases the ability to distinguish between arbitrary words), which is at the core of the LM objective and which is highly correlated with many of the other linguistic properties, such as POS. We argue that even if we do see a significant drop in performance with the control task, this says little on the validity of the results of removal of the linguistic property (e.g. POS).
However, for completeness, we provide the results in Figure \ref{fig:INLP_task_over_time_masked_hewitt_control}. As can be seen from this figure, this control’s slope is smaller than the one of the amnesic probing, suggesting that those directions have less behavioral influence. However, the slopes are steeper than the ‘Rand’ experiment. This is due to the identity removal of groups of words, due to the label shuffle, as suggested in their setup. This is the reason we believe this test is not adequate in our case, and why we provide other tests to control for our method: Rand and Selectivity (\S \ref{sec:method_controls}).

\end{document}